\documentclass[conference]{IEEEtran}
\usepackage{cite}
\usepackage[pdftex]{graphicx}
\usepackage[cmex10]{amsmath}
\usepackage{amssymb}

\usepackage{times}
\usepackage{array}
\usepackage[tight,footnotesize]{subfigure}
\usepackage[font=footnotesize]{subfig}
\usepackage{graphicx}
\usepackage{mathtools}
\usepackage{microtype}
\usepackage{amsmath}

\def\intersect{{\cap}}

\usepackage{alltt}
\usepackage{shadethm}
\newshadetheorem{comment}{Comment}

\def\boxx{{\vcenter{\vbox{\hrule height.3pt
          \hbox{\vrule width.3pt height6pt
          \kern6pt\vrule width.3pt}\hrule height.3pt}}\;}}

\def\impos{{\;\vcenter{\hbox{\rule{5mm}{0.2mm}}} \vcenter{\hbox{\rule{1.5mm}{1.5mm}}} \;}}

\def\lrarrow{\leftrightarrow \kern-8pt \rightarrow}

\def\2{\frac{1}{2}}

\def\beq{\begin{eqnarray}}
\def\eeq{\end{eqnarray}}
\def\2{\frac{1}{2}}
\def\AND{\,\wedge\,}

\newtheorem{example}{Example}

\newtheorem{definition}{Definition}

\def\lrarrow{\leftrightarrow \kern-8pt \rightarrow}

\def\frightarrow{\rightarrow \kern-11pt /~~}
\def\reducesto{\simeq \kern -3pt >}
\def\intersection{\cap}

\def\Server{{\rm Server}}
\def\Client{{\rm Client}}
\def\Proxy{{\rm Proxy}}

\usepackage{fancyhdr}					
\begin{document}
\newcommand{\strust}[1]{\stackrel{\tau:#1}{\longrightarrow}}
\newcommand{\trust}[1]{\stackrel{#1}{{\rm\bf ~Trusts~}}}
\newcommand{\promise}[1]{\xrightarrow{#1}}
\newcommand{\revpromise}[1]{\xleftarrow{#1} }
\newcommand{\assoc}[1]{{\xrightharpoondown{#1}} }
\newcommand{\rassoc}[1]{{\xleftharpoondown{#1}} }
\newcommand{\imposition}[1]{\stackrel{#1}{\impos}}
\newcommand{\scopepromise}[2]{\xrightarrow[#2]{#1}}
\newcommand{\handshake}[1]{\xleftrightarrow{#1} \kern-8pt \xrightarrow{} }
\newcommand{\cpromise}[1]{\stackrel{#1}{\frightarrow}}
\newcommand{\policy}{\stackrel{P}{\equiv}}
\newcommand{\field}[1]{\mathbf{#1}}
\newcommand{\bundle}[1]{\stackrel{#1}{\Longrightarrow}}

\title{Cooperation in Human and Machine Agents\\\large Promise Theory Considerations}

\author{Mark Burgess\\ChiTek-i AS\\12 April 2026}
\maketitle
\IEEEpeerreviewmaketitle

\renewcommand{\arraystretch}{1.4}

\begin{abstract}
  Agent based systems are more common than we may think.  A Promise
  Theory perspective on cooperation, in systems of human-machine
  agents, offers a unified perspective on organization and functional
  design with semi-automated efforts, in terms of the abstract
  properties of autonomous agents, This applies to human efforts,
  hardware systems, software, and artificial intelligence, with and
  without management. One may ask how does a reasoning system of
  components keep to an intended purpose?  As the agent paradigm is
  now being revived, in connection with artificial intelligence
  agents, I revisit established principles of agent cooperation, as applied to
  humans, machines, and their mutual interactions. Promise Theory
  represents the fundamentals of signalling, comprehension, trust,
  risk, and feedback between agents, and offers some lessons about
  success and failure.
\end{abstract}

\hyphenation{similarity}
\tableofcontents

\section{Introduction} 

Autonomy is the base state of any operational entity, human or
machine. Paradoxically, it is also an unconventional notion in
technology, which focuses historically on totally subordinated or
``managed'' devices, governed by ``command and control''.  That
idealized notion of control has long been viewed as the natural
`force for order' in a world of slowly varying, self-regulating
systems, but today it falls short when describing the multitude of
methods, at all scales of space and time, that render such direct
command infeasible.

Promise Theory\cite{burgessC1,burgessDSOM2005,promisebook} was
proposed in 2004 to address the issue of {\em autonomy} as a fundamental viewpoint
in software `robot' systems, specifically CFEngine\cite{burgessC1}. It was a response to the perceived lack of
progress in the field of multi-agent distributed systems, which had
foundered on universally adopting logical deontic
reasoning\cite{marriott1,deontic,deonticcontracts} and control
theory\cite{controltheory,burgessC4,diaoDSOM2002} as a model of
cybernetic control.  Autonomous agents are significantly different
from remotely managed robotic entities; they must handle every
scenario independently; they are intrinsically non-deterministic, and
they require more comprehensive specifications to constrain their
behaviours.  Promise Theory seeks to address these issues.

Trying to apply (deontic) obligation logics, together with ideas from
Control Theory, led to unrealistic remote command models of behaviour
that could not be scaled beyond simple state
controllers.  The deployment of autonomous agent CFEngine (1993-) as a
datacentre management engine, around the world, was a watershed moment that proved
autonomous agents were indeed possible and could be run on a massive
scale, subject to certain self-maintaining
policies\cite{burgessC11}. CFEngine employed {\em autonomous} agents, and marked
the starting point for Promise Theory, in terms of
limitations and capabilities of agents working alone or in groups.

Autonomous agents differ from fully subordinated, remotely controlled
automata, because---when conflicts arise between multiple
agents---they cannot be resolved simply by shouting louder. Autonomous
agents do not have to pay attention; they must have the capacity to
adapt to ambient conditions of their own volition, form structures (like swarms), and solve
challenges through the balance of offer and avoidance.

The basic theory of cooperation has already been elaborated in
\cite{siriAIMS1,spacetime2}. However, since this was long before the
current wave of agent language models, related to artificial
intelligence, took hold, it is appropriate to review and update the
basic findings of Promise Theory as a guide to both strategy and
governance.
There has been a renewed interest in Promise Theory, in connection
with Artificial Intelligence, especially Language Models, and
robotics, since it has already formulated many of the issues authors
confront when deploying systems.
Here, I shall try to present the results in a general way,
without referring to specific implementations.

What Promise Theory emphasizes is that functional {\em roles} matter
more than technological will, as long as one is honest in
characterizing them. Roles are simply expressions of particular
promises, which in turn are a declaration of intent that encompasses
two aspects: what is being promised (a qualitative or semantic issue)
and how much and how often (a quantitative or dynamic issue).

The plan for this document is as follows. Following a brief recap
of Promise Theory basic ideas and notations, I discuss the notion of
voluntary cooperation and subordination for autonomous agents, along with
the all-important Downstream Principle. Next, we consider how shared language semantics
are a prerequisite for communicating intent between independent agents,
and how this issue alone leads to significant uncertainty both for humans and machines.
Then, we address the dynamic {\em quantitative} issue of trust its relation to
an agent's risk appetite. Finally, I bring together the points to
describe the aims and pitfalls of cooperating to enact collective action and
explain the role of fixed points in defining idealized attractor states,


\bigskip
\section{Basic Promise Theory}
\bigskip
The fundamental tenet of Promise Theory is this:
\bigskip
\begin{quote}
{\em ``No agent may promise
  anything on behalf of any agent but itself.''}
\end{quote}
\bigskip
Take a moment to let this sink in, because attempts to work around
this causal limitation account for almost all misunderstandings and errors in agent
systems. Working through the logical consequences of this statement is
what Promise Theory is about, in its entirety. The remainder of this paper
is about unpicking the detailed consequences of reining in autonomous behaviour.

Autonomy means that agents stand alone, as sovereigns in command of
their own interior resources.  The concept of `agency' is sometimes
confused with an idea about free will, but this is false. Agency
simply means that there may exist entities for which no outside
influence can fully determine their behaviour. Some directed effort can come from within
a bounded region that we call an agent, and this only requires
there to be some internal modification of state.

This is a deeply unnatural idea for humans to accept, both on a
personal as well as an engineering level.  The desire for control is
deeply ingrained into us, perhaps because we have evolved to make
extensive use of our hands.  We want to touch, manipulate, wrestle
with adversaries, push buttons, and pull levers; but, if we are
suddenly unable to touch and are forced to work in a different way,
what could we do? How does the world look to a single cell, to a tree,
or talking to someone on the other end of a telephone line?

In Promise Theory, every system is modelled as a collection of agents, of different varieties,
denoted $A_i$, whose properties remain to be described.  Agents may be
atoms, cells, humans, or machines, hardware or software, etc. They may
be any shape or size (if that has any meaning). Agents may contain
other agents by composition.  What happens outside agents is not
Promise Theory's concern: we focus on how agents influence one
another, leading to expected outcomes.

Crucially, agents are finite but have some measure of 'interior'
resources, depending on their nature.  Their capabilities, energy, and
other resources are all finite. Agents are thus a model for any bounded {\em active
  process}\cite{promisebook}.
\begin{itemize}
\item The interior of an agent is not observable to other agents.
\item Agents may possess or form intentional behaviours.
\item An intention is a causally directed notion about a possible outcome.
  The number and scope of possible outcomes depends
on whatever capabilities the agent has. 

\item  On their exterior, agents can communicate with other agents to express
their intentions. The sharing of an intention is what we mean by a
promise.
\end{itemize}
None of these capabilities represent intelligent thought or free will, they are
simply process characteristics.

\bigskip
\begin{example}[APIs and asynchonrous interfaces]
  At this stage, many programmers will identify promises they know as
  Application Programmer Interfaces (APIs). API functions are
  certainly examples of promises offered by software agents. In
  languages that offer asynchronous responses, e.g.  JavaScript, there
  are features branded as `promises', which are a particular case
  compatible with the Promise Theory notion of a promise.  The
  molecular donor and receptor spikes on cells are similarly effective
  promises made by cells, and door handles and emergency exit signs
  are also agents that make promises.
\end{example}
\bigskip

Promise Theory deals with general notions, not only
information technology concepts. In doing so, it offers an opportunity to
treat humans and other forms of machinery on a common basis---to view technology in the
context of a human society.

\subsection{Notation and definitions}

An abstract notation for promises is helpful for formulating problems
more clearly.
The generic label for agents in Promise Theory is $A_i$, where Latin
subscripts $i,j,k,\ldots$ numbers distinguishable agents for
convenience (these effectively become coordinates for the agents).  We shall often
use the symbols $S_i$ and $R_j$, instead, for agents to emphasize
their roles as source (initiator) and receiver (reactor).
So the schematic flow of reasoning is:
\begin{enumerate}
\item $S$ offers (+ promises) data.
\item $R$ accepts (- promises) or rejects the data, either in full or in part.
\item $R$ observes and forms an assessment $\alpha_R(.)$ of what it receives.  
\end{enumerate}
Observations and assessments are limited only by the capabilities of the agent.
They may be economic assessments, to minimize or maximize some parameter. They may be
semantic assessments to stay close to a logical or symbolic constraint.
They are a key culprit in the indeterminism of agent systems.
One
should be careful in anthropomorphizing too much. One uses the words promise, assessment, etc
because the convey the correct meaning in the limiting case of complex human agents,
but equally assumptions have to be scaled back when being used for simpler agents with limited
capabilities. For a molecule, an acceptance of a promise may be equivalent to
accepting a donor molecule by docking with it.

The elementary agents of a system embody {\em processes}, any of which
may express promises about state and services. Processes are hosted at agent locations 
$A_i,S_i,R_i$ etc.
Agents have finite resources, so their capabilities are limited. However, they might
over-promise accidentally or deliberately, e.g. promise 10 parking spaces for 20 cars,
or 100 seats for 150 passengers. Such a promise can be resolved by time-sharing.

Promises may refer to agents, e.g.
\begin{itemize}
\item A scalar promise refers to no other agents, e.g. I promise to brush my teeth $+b$.
\item A vector promise refers to a single third party, e.g. I promise to follow orders from $A_3$: $=b(A_3)$.
\item A tensor promise may refer to any number of agents, e.g. I promise to get the best deal from
suppliers $+b(A_1,A_2,\ldots,A_n)$.
\end{itemize}
Promise also come in two major flavours.
An {\em offer} promise, written with body $+b_i$ made by $S_i$ to
$R_j$ is written:
\beq
A_i\promise{+b_i} A_j,
\eeq
where the $+$ refers to an offer of some information or behaviour (e.g. a service).
This is a part of $A_i$'s autonomous behaviour, and the promise constrains only $A_i$.

$A_j$ may or may not accept this offer, by making a dual {\em acceptance} promise, marked $-b$ to denote the
orientation of intent:
\beq
A_j\promise{-b_j} A_i.
\eeq
Only if {\em both} offer and acceptance promises are kept can an influence be expected to
pass from $A_i$ to $A_j$. In general, the offer and acceptance may not match
precisely, in which case the propagated information will be the overlap (mutual information)
\beq
b_\intersect = b_i \intersection b_j,
\eeq
in the manner of mutual information\cite{shannon1,cover1}.

What form does the abstract $\pm b$ take? A form of signalling between
them may take the form of a language, from regular language all the way up to
advanced (human) language\cite{lewis1}.

Agents can offer and accept promised outcomes, so they can also make promises conditionally.
To promise $b$ on receipt of a certain conditional $c$ (e.g. if the book arrives, I will read it by Monday):
\beq
A_i \promise{+b | c} A_j.
\eeq
This is not yet a promise, without also passing information
about the condition. The receipt of $c$ from some other agent,
\beq
A_k \promise{+c} A_i
\eeq
must be taken together with the acceptance of $c$ and signalling of the acceptance to $A_j$
\beq
A_i &\promise{-c}& A_k\nonumber\\
A_i &\promise{-c}& A_j
\eeq

Such conditionality is subtle and potentially misleading, as it involves agents
as `middle men', which introduces non-deterministic aspects. Agents
are not reliable relays in the sense one might expect.


\bigskip
\section{Behaviour as voluntary cooperation}

The simplest building block
of promise theory is a single promise exchange.
It's convenient to base the discussion around this concrete binding.
Consider the two promises between two agents:
\beq
\pi_S: S &\promise{+b_S}& R\nonumber\\
\pi_R: R &\promise{-b_R}& S,\label{SR}
\eeq
representing the offer and acceptance of a single
intention $b_S$, about the behaviour of the originator agent $S$ (the sender) to $R$
(the receiver).  This is
referred to as an example of {\em voluntary cooperation}, because there is
nothing forcing $S$ to offer $b_S$ to $R$. Similarly, $R$ is not under
any obligation to accept whatever it is that $S$ is offering.  This is
to be contrasted with the notion of an {\em imposition}: \beq I_S: S &\imposition{+b_S}&
R, \eeq which is an attempt to induce the acceptance of $b_S$ by
force. Impositions are a generally ineffective way of seeking cooperation,
because it's unclear whether the receiver is even aware of the
imposition, and further whether it has any reason to comply. Imagine throwing a ball to someone without arranging it
first\footnote{These two cases correspond to the widely known TCP and
  UDP Internet protocols. TCP is a promise, while UDP is an
  imposition.}.

Many technologists will regard this coordination
issue as a management issue.  If a receiver wasn't listening, surely
we just kick it to restart the service?  For a single promise this viewpoint is a plausible
in some cases, if a little simplistic, as an
answer for information technology. It doesn't apply to biological agents or organizations.
More generally, a missed signal can change the
entire evolution of the system in a complex network of time-dependent
interactions.

\subsection{The receiver wins: the Downstream Principle}

One key consequence of autonomy is the inversion of how we think about
causality.
In classical science, causality is an arrow from now to future.
In technology, we are used to exploiting this for commanding and controlling ``hands on''.
The source of a signal is expected to be the origin of causation: the authoritative
decision point.

Autonomy refines this, by breaking it into two parts: the
authoritative source of message may belong to an `upstream' source agent, but the
autonomy of the `downstream' receiver means that it is under no obligation to
accept the message. Thus, the decision point, for determining the exchange of a cooperative signal, is
weighted ultimately in favour of the receiver. A downstream agent can always reject or even
completely alter an outcome downstream. The practical outcome of this is that the role of
causation is turned upside down for autonomous agents. Agents can see what they want to see,
and thus may distort the ``truth'' of the system.

The significance of autonomy is thus that every interaction is made entirely at the
pleasure of the {\em receiver}.  This is another confusing idea to traditional
technologists, who are used to viewing the source as the authoritative
signal. A source, however, is only as good as its signal can be heard
downstream. Thus, Promise Theory recognizes the {\em Downstream
  Principle}.

\bigskip
\begin{definition}[Downstream Principle]
  Agents downstream, i.e. on the receiving end of an offered promise, have the ultimate
  power of decision over the outcome.
\end{definition}
\bigskip
If a provider fails to keep a promise, the downstream agent only has its
own policy to blame. It could or should have sourced more than one
provider, planned for the promise not keeping kept, and sought out
redundancy from multiple sources. In general, responsibility for
outcomes flows downstream.

\subsection{Assessment of responsibility}

In a service interaction, the client user of a system is downstream of everything the system provides,
so that user is responsible for their own use of the system: it doesn't help to blame the newspaper for receiving bad news,
or the chatbot for a bad answer. One takes on the risk of an agent's promise not being
kept when choosing to engage with it. I'll return to this below, in connection with trust.

With only two agents, this is easy to grasp. In 
a distributed system with multiple stakeholders and a complex network of dependencies, it
may be much harder. We can try to state the problem:

\bigskip
\begin{definition}[Responsibility for keeping a promise]

\begin{itemize}
\item {\em Causal responsibility}: When an agent relies on a promise
  from another, as a dependency, in order to keep its own conditional
  promise, the causal responsibility it has refers to the freedom to
  obtain the promised service elsewhere: from its own autonomous
  choice of partner in the presence of redundant alternatives.

\item {\em Moral responsibility} (culpability) is a human assessment,
  about whether agent outcomes stem from good or for bad intent, hence
  it cannot be formalized except as a norm or in law. This is not a
systemic issue, only a subjective assessment.
\end{itemize}
\end{definition}
\bigskip

When an agent has no choice but to rely on an upstream promise, it has
effectively promised away its own independence.
In other words, a client user is always responsible for its own use of a service, because it
was not forced to do so. If the service
does not keep its promise, this is unfortunate but also to be expected, because guarantees
are impossible, and it is the autonomous responsibility of the user to allow for that.
Thus the system, in total, may be said to have flaws by design.

\bigskip
\section{Communicating intended outcomes}

In order to cooperate, agents have to signal their intentions somehow.
This might be a simple or a complex effort, depending on the kind of agent.
Promises cannot drive expectations without a shared language for signalling intent.
Between any two agents, there are three languages to consider:
\begin{itemize}
 \item The language of the sender,
 \item The language of the receiver, and
   \item The exchange co-language of cooperation between the two.
\end{itemize}
An intention, e.g. $b_S$, is the basic formulation of a `course of behaviour' we
would like to follow, so we need to discuss what language it speaks.

\subsection{Shared language}

Consider a set of domain specific languages for signalling intent between agents: $\ell^{(\alpha)}$, where $\alpha$ denotes
each distinct language. A language will consist
of a vocabulary
\beq
\ell^{(\alpha)} \rightarrow \{ \beta_a^{(\alpha)} \},
\eeq
i.e. words that combine somehow to signal intent. Such a language $\ell^{(\alpha)}$
is the language of promise bodies emanating from some agent $A_i$:
\beq
b(A_i) = \sum_{a=1}^{C^{(\alpha)}} \; c_a\,\beta_a(A_i),\label{linear}
\eeq
where $C^{(\alpha)} = \dim(\{\beta^{(\alpha)}_a\})$ is the dimension of the vocabulary.

\begin{figure}[ht]
\begin{center}
\includegraphics[width=5cm]{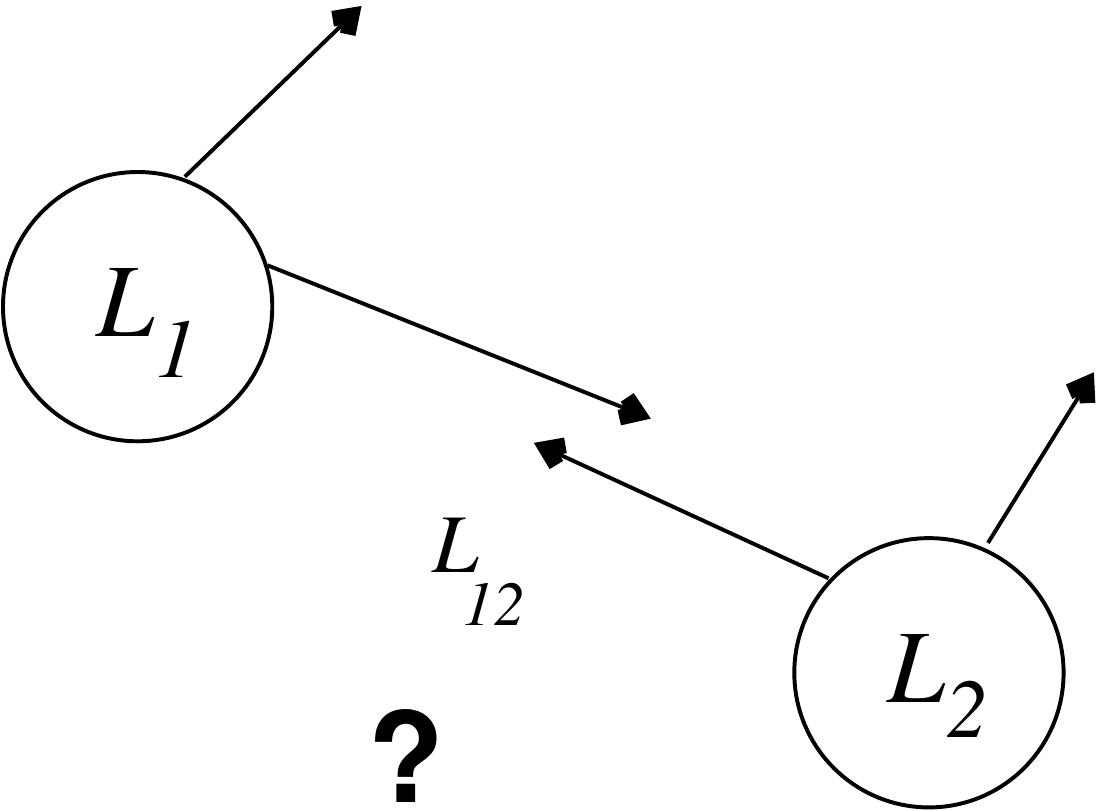}
\caption{\small Autonomous agents each have their own internal language.
  There is no authority that calibrates these to be the same without the agents' consent.
  This means that, between any two agents there is a subset of language that overlaps
  and is comprehensible to both, but may still mean different things to each. What
  happens where these meet?\label{lang}}
\end{center}
\end{figure}

In order for languages to be mutually intelligible, they must be translatable. For simple
low order languages, such a translation might be a linear equivalence ``word for word'',
\beq
b(A_i) = \sum_{a=1}^{C^{(\alpha)}} \; c_a\,\beta_a(A_i) = \sum_{b=1}^{C^{(\alpha')}} \; c_b\,\beta_b(A_j),
\eeq
where $\alpha=\alpha'$. In general, however, one language may need more words in order to
explain the meaning in a satisfactory way.

The larger the dimension of the languages, the greater the probability that they will be able to
overlap. If they share common words,
or at least have equivalent words, they may be translatable for some intention or utterance.
A small language can assign a very precise meaning to a word, that will not be reused outside of a controlled
context, but it can't say many things.
Thus, there is a trade-off
between trying to lock down meaning in a small language, and allowing free
expression over a broad language of exchange. This is the trade-off one faces in technology
systems today. 

\begin{itemize}
\item If we rely on small signal sets, the meanings can be precise to the agent
  signalling but may be misunderstood by the receiver, unless they have calibrated their
  interpretations to a specific version of a specific language.

  This is the approach used in engineering signalling protocols, especially for machines and for military systems.

\item If we have a broad vocabulary, we might be able to explain concepts in detail to avoid
  misunderstandings or to compensate for missing words, but this is not a guarantee. Agents may have to
  talk their way to a calibration of meaning for some time to stabilize their relationship.
\end{itemize}

\bigskip
\begin{example}[Home assistants]
  The way language comprehension is applied is quite uneven in
  contemporary tech.  Speech to text conversion is quite advanced, and
  text to text language translation is also quite good for many
  purposes, yet home assistants like Google Home, Alexa, Siri, etc,
  are not very clever at understanding speech.  They accept a very
  simple set of instructions and any deviation from these leads to
  incomprehension. ``Turn the light off and close the blinds'' ought
  to be quite straightforward and yet flummoxes a machine designed only
  to respond to simple monotonic imperatives.
\end{example}
\bigskip

For humans, who have come to rely on large languages, the discipline
of small language expression is often difficult.  In a higher order
languages, with large vocabularies, words may be repeated several
times in the same string, and may be combined with others in order to
form new concepts; then there is no simple linear decomposition: there
is only an ordered sequence of symbols, like a line integral over a
field.  In the transition from large to small, one may select a subset
of a full language so that a small set can masquerade as full
language. This presents a risk to interpretation.  Consider the
language of mathematics, for instance, which uses a subset of English
with special jargon so that common meanings are not always
honoured. In music, `rap' artists often use words with rhyming
associations involving levels of indirect association to convey
meaning so someone who has a similar background.

A unit of intent, in some language, is now represented as an ordered sequence or path
over words in the language, which we can denote by $L_{SR}$ from an agent using $L_S$ internally
to an agent using $L_R$ internally. The co-language understood by both
\beq
b_S(L_S) \;\intersect\; b_R(L_R) \in L_{SR}.
\eeq
Agents may need to adapt their signalling in order to be understood, and there is nothing
that guarantees that intent will be transmitted faithfully, since autonomous agents may have
nothing to align them. We can write this schematically as a quasi-functional path integral through the
effective space of intentions, to illustrate the semantics and make contact with representations in the mathematics
of fields: a language element is a path dependent sequence of elements, contextualized by
some ambient context $c(p)$ along the intended path:
\beq
L \mapsto P \int_B^E dI \;c(I) \;\beta(I).
\eeq
where $\beta(m)$ may depend on the position in the message, or the context, and $B$ and $E$ are the
beginning and end of the fragment.

This traversal of ideas in a `semantic spacetime' is one way to form
local knowledge representations\cite{spacetime3} and is related to
discussions about knowledge maps, and graphs in the Semantic
Web\cite{rdf,robertson1}. Such knowledge representations suffer from
the same problem as their base languages: they do not necessarily
match in their intended meanings, because they are structured by some
curator's idea of how things play out. This is the ontology
problem\cite{rasmussen1,autonomic_ont}.  Ontologies work by trying to
decompose language into isolated concepts that are unique and
separable within some hierarchy of ``types'', and are then arranged in
some connected `spanning tree', which promises greater significance
than can be attributed to it in a fair fight. Mapping and ontology
are, in fact, an agent strategy.  In practice, such type-inspired
decompositions are as non-unique as languages, and yet they offer less
expressibility because they have no grammar for composition. So what
one gains, in making clear promises between the concepts, is lost by
placing unrealistic constraints for the sake of differentiation.  This
is simply a step backwards from a large and complex language to a
smaller simplistic one, with a restricted vocabulary of assigned
meanings, that now needs a versioned calibration to match reality.

\subsection{Consistency and comprehension of language}

Certain languages, which can promise both the {\em doing} and the {\em undoing}, i.e. the reversal of some outcome,
are useful for enabling self-assessment and consistency of meaning across multiple exchanges.
Did I do what I think I did? Did you agree that I did what I think I did?
Such languages ensure that each promise has a meaningful and unique inverse.
In order for two languages to have consistent intent
we must have an intention expressed in language $\alpha$ should have the same meaning
in language $\beta$, to the best effort:
\beq
b^{\alpha} = L_{\alpha\beta} (b^{\beta}),
\eeq
so that translating back to the original has no effect
\beq
b^{\beta} = L_{\beta\alpha}(b^{\alpha}) = L_{\beta\alpha}(L_{\alpha\beta} (b^{\beta})),
\eeq
or the reversals are true inverses
\beq
L_{\beta\alpha}\cdot L_{\alpha\beta} = I,
\eeq
which means that such a language $L_{\alpha\beta}$ would be unitary.
This is a very strong constraint, which is unlikely to be achievable
between independent agents without highly stringent
accommodations. Thus, in practice, agents should expect to
misunderstand one another's intentions to some level and thus make
additional promises to accommodate such misunderstandings, when
cooperating\footnote{In restricted languages, Shannon proved that
  formal error correction can mitigate poor
  transmission\cite{shannon1,cover1}, but not intentional
  misinterpretation}.

Suppose there exist a few  multi-valued elements (words with multiple
meanings), creating ambiguity of interpretation, then a language
clearly cannot be invertible. One way to handle this would be to not
rely on single words, but rather expand discourse into unique sentences on a
larger scale of explanation. Thus one can approximate an effective unitarity
of communication by expanding extent (decompression)\footnote{Readers should not misunderstand
  this intentionally: an agent does not have a choice about
  intentionally misinterpreting a message. It projects its own intent
  to the best of its ability, without presupposing what its own
  intentions might be. It takes only a small misalignment to end up
  with a deviation from what was intended by the sender. All agents
  can act in good faith and still end up with an error.}. This is a risky
strategy, because additional words may introduce inconsistency or contradictions
as well as clarifications. Again, there is a trade-off between restriction on
intent and restriction on constraint language.

Finally, language is the important medium for knowledge representation
when saving insights for the future or sharing with.  As versions of
language evolve, stored information may become unreadable or
unintelligible.  It can also be misunderstood because the unspoken
dependency of presumed context, assumed to be ``common knowledge'' is
now missing. If no agent has a clear understanding of the context in
which memorized input was relevant, then it becomes a liability.

\subsection{Comprehension uncertainty}

When we look for keywords, such as directives or signals, in a piece of text, we are extracting
intent from the text. The phrases that trigger promises may be considered 'possible intentions', i.e.
directions in which one could possibly choose to go, or tasks one might undertake.

One way to maintain coherence is to limit the common language of by providing a finite menu
or directory of cases. This is how a fast food outlet can perform efficiently and mechanically,
as compared to a diner in which can order breakfast by personal specification.

When the multiplicity of intentions grows large, perhaps forming a quasi continuum (as in human
systems), behaviours lose their coherence. The machine-like qualities of robustness and reliability
become washed out and effectively 'unintentional' as signal turns to noise.

\bigskip
\begin{example}[Simple signal language]
  
  Consider a body language with alphabet
  \beq
  \beta = \{ \text{\sc send, receive, seek, forward, back} \}
  \eeq
  and
  \beq
  \beta' = \{ \text{\sc put, get, append} \}
  \eeq

  then we can translate these, referring to (\ref{linear}):

\beq
c_1'\beta_1' = \text{\sc put} &=& c_1\beta_1 = \text{\sc send}\\
c_2'\beta_2' = \text{\sc get} &=&  c_2\beta_2 = \text{\sc receive}\\
c_3'\beta_3' = \text{\sc append} &=& c_2'\beta_2' + c_4'\beta_4'+c_1'\beta_1' \nonumber\\
&=& \text{\sc seek+forward+send}\nonumber\\
\eeq
There is a translation matrix:
\beq
L_{a'a} = \left( 
\begin{array}{ccccc}
1 & 0 & 0 & 0 & 0\\
0 & 1 & 0 & 0 & 0\\
1 & 0 & 1 & 1 & 0\\
\end{array}
\right)
\eeq
which, in this case, is not invertible. Hence the language is translatable in one direction
only.
\end{example}
\bigskip

In principle, however, it should be possible to restrict $\ell^{(\alpha)}$, such that translation may be
performed faithfully as a bijection, by postulating a `table of elements' for the chemistry
of all promises in a semantic spacetime:
\beq
\ell^{(a)} \leftrightarrow \ell^{(b)} ~~~ \forall \;a,b \in {\cal L}.
\eeq

\begin{figure}[ht]
\begin{center}
\includegraphics[width=7.5cm]{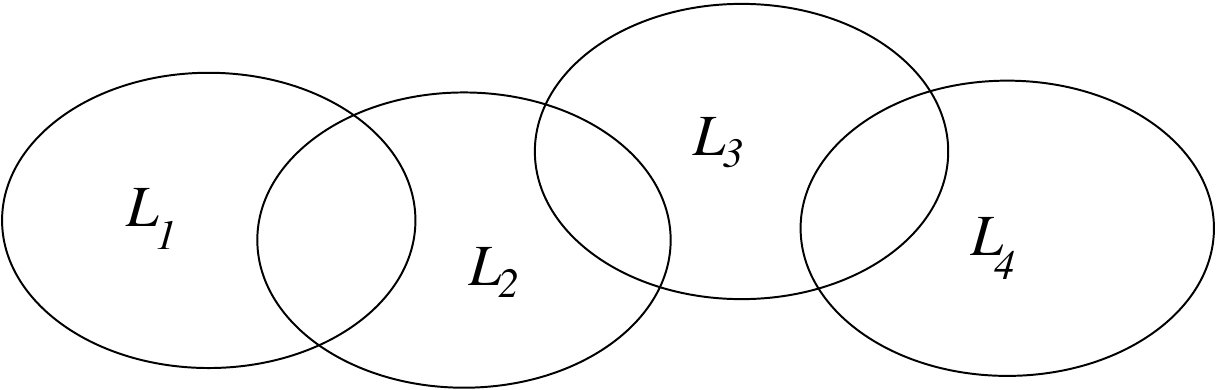}
\caption{\small Overlapping patches of language require all agents in a patch
to use a compatible language, and for (at least some) agents in each patch
to comprehend a faithful translation of a neighbouring language, in order to
bring long range order.\label{languages}}
\end{center}
\end{figure}

It is not only the language in its entirety that needs to be compatible, but the way each agent
uses the language. When dealing with large language models, there is a temptation to believe that
agents sharing the same model must automatically understand one another. However, this need not be
the case, given that each picks a selection based on its own context (which is autonomous and independent).

\begin{example}[Small vs large]
  A small language with a selection of words with standard meanings is
  only standard to its creator. If the receiver doesn't comprehend
  what is meant, repeating the same words doesn't make it clearer. In
  a larger language, the sender can go on in the hope that eventually
  the right intention will be communicated. However, at some point
  saying too much could make things worse. Agents can never know when
  they have reached the optimum without actual a (promise) dialogue and
  mutual assessment.
\end{example}

Autonomous agents are never certain.

\subsection{Language uncertainty during serial-parallel composition}

The impact of language comprehension at junctions, involving a
composition of agents into larger structures, where promises depend on
one another will multiply as a systemic fault, following the
conventional algebra of fault dependencies\cite{treatise2}.

\begin{figure}[ht]
\begin{center}
\includegraphics[width=7cm]{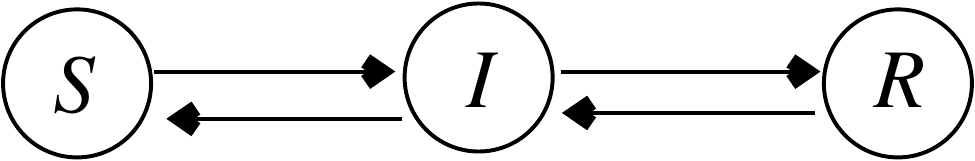}
\caption{\small Cooperation in parallel and in series presents different challenges, each with
  their own uncertainty.\label{seripara}}
\end{center}
\end{figure}

\bigskip
\begin{example}[Component assembly]
  For instance, a radio is a superagent formed from
smaller agents (resistors, capacitors, diodes, etc each of which makes
a simple promise); their composition into a circuit is what enables
the promise of radio sound.  Promises that depend on other agents may
be combined serially, or aggregated from parallel upstream promises
(see figure \ref{seripara}).
\end{example}
\bigskip

\begin{figure}[ht]
\begin{center}
\includegraphics[width=6cm]{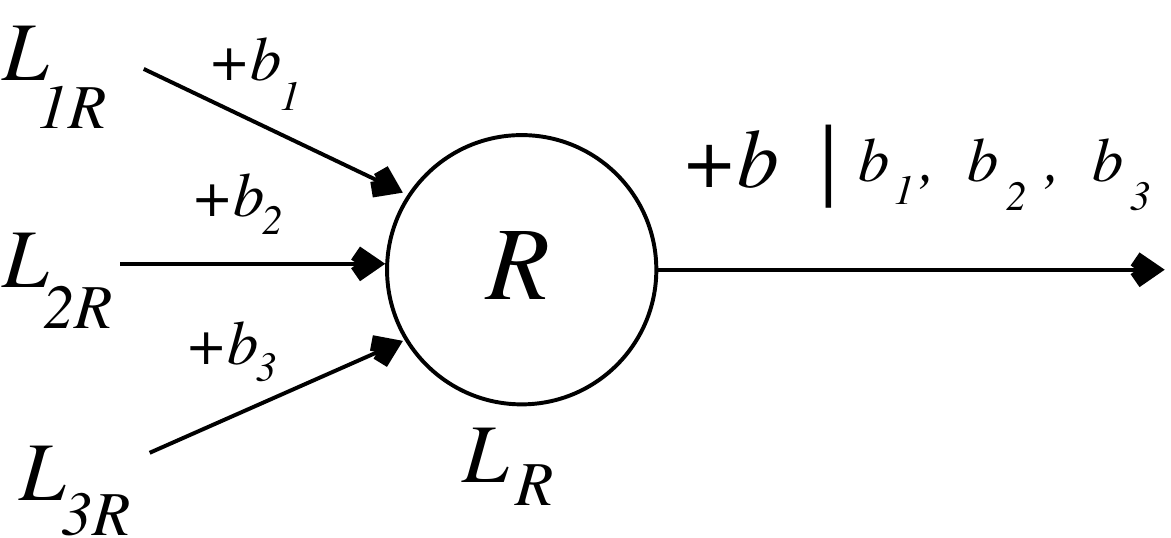}
\caption{\small The aggregation of promises to assemble a new (tensor) promise is dependent on the overlap between each pair of agents, which involves language uncertainty $L_{iR}$ and measure uncertainty $b^{(+)}_i\intersect b^{(-)}_{Ri}$.\label{aggregate}}
\end{center}
\end{figure}

These configurations each lead to their own uncertainties. Figure figure \ref{aggregate} combines a number of
inputs into a single promise dependent on all. If the inputs are redundant sources, then the dependent promise
has a resilience. However, if each is a different independent input, such as data from a unique sensor, then
the output becomes fragile in proportion to the promise that aggregates them.

\beq
S_i &\promise{+b_i}& R\nonumber\\
R &\promise{-b_i'}& S_i\nonumber\\
R &\promise{b \;|\; b_1, b_2, b_3 \ldots }& ?
\eeq

Each agent forms its autonomous assessment of the intention conveyed
in a received promise, and replaces it entirely with its own
promise. Ultimately, it cannot be regulated or controlled without
establishing an agreed configuration in advance. For instance, when a
collection of agents is composed into a new superagent (such as a
radio from transistors, resistors etc), then there is a master agent
providing a calibrating design for the construction. The participating
component agents basically agree to subordinate themselves to this
design. In a human organization, one attempts to replicate this into a
collaborative workflow, but humans (with their many concerns and
distractions) may change their minds or fail to keep rigid promises
required for such predictability. Generally, the more complex or high
the level of the agent and its interior processes, the less
predictable and reliable a collaboration will be\footnote{Daniel
  Mezick remarks on an aphorism by Stuart Kauffman: `Work is the
  release of energy through a few degrees of freedom', i.e. there are
  enabling constraints (an explosion through a gun barrel turns an
  undirected energy release into an intentional force). Constraint is,
  in fact, the architect of intention, and language is its language!}.

In the serial case of figure \ref{seripara}, which represents a relay station or supply chain:
\beq
S &\promise{+b_S}& I\nonumber\\
I &\promise{-b_{S'}}& S\nonumber\\
I &\promise{+b_{S''} \; |\;b_{S'}}& R\nonumber\\
I &\promise{+b_{S'}}& R\nonumber\\
R &\promise{-b_{S''}}& I
\eeq
The intermediate agent becomes the chief suspect when transmission of intent from $S$ to $R$ fails.
How can we interpret the promise body sets: $b_S = b_{S'} = b_{S''}$?
An intermediate agent can always inject or distort a message, either deliberately or unwittingly.
The language problem is one aspect of this; timing delays and sampling distortions are another.
What can we expect of an ensemble of agents interacting.
\begin{itemize}
\item The potential inhomogeneity of the agents (in both capability and comprehension) is in play. Do they share a common language and will they understand the same words in the same way?
\item Autonomy implies that any agent can inject its own intentions at any stage, so we cannot automatically assume that an autonomous agent is a faithful servant.
\end{itemize}

\bigskip
\begin{example}[Law and ontology]
  Limiting of vocabulary to avoid uncertainty is what lawyers and
  ontologists try to do.  This is a difficult job, requiring much
  effort, so many human agents will try to avoid it and improvise
  their way to a resolution, building on `stigmeric' or commonly
  established norms (see below), rather than specific promises.  Creating a limited
  vocabulary language has the further impact of making something
  domain specific masquerading as something ordinary.
\end{example}
\bigskip

\subsection{Autonomous agents appear non-deterministic}

Even if agents act deterministically, within their interiors, the outcomes of
their promised actions, as measured by others must occur
{\em non-deterministically}. This is a consequence of the Nyquist-Shannon sampling law\cite{shannon1,cover1},
and we shall use it below in connection with trust and risk.

Consider, once again, the promises in (\ref{SR}).
The rate at which an agent can listen for and receive messages, in order to respond, depends
its interior capabilities and resources. This becomes important when
discussing trust, since it is the fundamental limitation that affects the way agents
perceive one another.
By the sampling
law\cite{shannon1,cover1}
\begin{quote}
\it ``If a function $x ( t )$ contains no frequencies higher than $B$,
then it can be completely determined from its sampled values at a sequence of points spaced less than
$1 / ( 2 B )$ time intervals apart.''
\end{quote}
Thus, in order to assess the promise from $S$, with perfect fidelity,
$R$ has to sample (i.e. be paying attention) at twice the rate at which $S$ may send.
Again, for a single promise, this is plausible. However, in a larger network,
it becomes paradoxical.
If all the
agents were to be similar in capability, then every agent would have
to sample twice as fast as every other, which is impossible. Conversely, if
they all sample at their own rate, there is necessarily a finite chance of missing a signal or
or misunderstanding a message.  Error correction can be applied to the
message, to restore its integrity, only if the sampling frequency is
already adequate and may result in a delay.

What appears to be one agent may be resolved into several independent
agents working together. An agent that appears to keep a number of different
and unrelated promises might actually be inferred to consist of a number of independent
agents.
Autonomous agents have no a priori homogeneity. Homogeneity, in general, arises only by
calibration to a single calibration source, or ``standard candle''.

\subsection{Intended purpose, free will}

A note is warranted, concerning use of the terms `intended meaning' or
`intended purpose', which for some readers will naturally bring up the
contentious philosophical issue of `free will'.  Some believe that
free will is a quality reserved specifically for humans, and that intentionality is
related to questions about consciousness.  To begin with, we can try to define these terms more clearly:
\begin{definition}[Intent]
  A selection of a single course of action, whose outcome satisfies certain constraints,  amongst a number of possible alternatives.
\end{definition}

\begin{definition}[Free will]
  The ability for an agent to make its own selection from a number of self-assembled
  alternatives, without interference from other agents.
\end{definition}

\begin{definition}[Consciousness]
  The ability for an agent to assess and reason about its own state along side the state of its environment
  and compose a catalogue of possible alternatives.
\end{definition}
\bigskip

Notice that free will doesn't claim that agents cannot listen to each other's input.
The only mysterious part of this is where a catalogue of possible alternatives may come from and whether
it is fully deterministic and predictable. Promise Theory reveals that a system composed of agents can never
be deterministic, because of finite sampling.

Intent (which has an independent meaning to consciousness, but which
may rely on it) relates to a cognitive awareness of an agent's
situation and the assessment of possible options. Intent is thus
related to an agent's `direction of travel' in some space of
possibility\footnote{For elementary machines, intent is simply a proxy
  for some human's intent. However, this view quickly becomes
  cumbersome.  A table, once fashioned, promises to hold items and
  form a surface for arranging other items.  A door handle promises to
  open and close a door. These promises remain true even after the
  original inventor is long gone.}.

The same is true for any entity that makes a dynamic choice.  The
extent to which an agent acts fully autonomously, or with complete freedom,
varies enormously with it circumstances.  In a sense, the possibilities
for future choice are all present in an agent's environment and
internal state. The selection of a specific choice may be
deterministically made, but nevertheless appear mysterious and free of
influence just as long as we do not see the process by which the
selection was made.  As long as intentions come from within, in the final instance, they
appear to be autonomously and freely made. This is true for humans
too. The extent to which an agent can think or make decisions freely
is a an elaborate illusion, shrouded in complexity.

It should not surprise us that simple agent mechanisms can execute
processes that make pre-existing selections, e.g. the automatic
transmission (gearbox) in a car can select from a menu of gear ratios in a
static environment, and even be cajoled into changing gears in
response to a human tweaking its sensors.  Going further, an agent
with much more sophisticated information, within a significantly more
complex environment with stigmergic trails from earlier generations
leading it in different directions, can easily select opportunities
that are latent within that dynamical environment, giving the
appearance of magical innovation.  The self assessment of intentions
leads to this kind of surprise: a system may be ripe with the (+)
promise of possibilities, and yet we only pay attention (-) to a small
part of them, due to our cognitive budgeting\cite{burgess_attention}.

For agents with `emotional' judgements, i.e. the ability to form quick and dirty `low information' responses, which
attract or repel the agent from certain scenarios, a functional `intent' already has a hardwired component.
Emotional signalling is an effective strategy for fast reasoning\cite{kahneman} and plays a key role
in human reasoning selection. Emotional communication is, like body language,
an additional dimension to expression that machines fail to grasp (as do some humans).

The deeper question of whether an agent could alter a purpose that was
assigned to it is then something like the question: can someone take
on a new job? If all its possible intentions are pathways to possible
existing outcomes, but not all are clear in its attention, then this
is really a question of how to allocate a cognitive budget to navigate
those unexplored avenues. An agent that spends time introverted in
self-examination will discover more of those opportunities than one
that spends most of its budget responding to external stimuli.  The
decision to select an `non-programmed action' is not in question. The
`free choice' is simply a selection from a largely unexplored menu of
pathways, initiated from within.

Assuming we can agree sufficiently to shape norms for the use of artificial agents,
policing the future will be an interesting challenge. One can only enforce laws if
a majority basically already believes in them.

What we mean by `intentionality' is the ability of an agent to select
from alternative next steps, thus aligning an interior process to an
outcome in some semantic representation.  The more complex the
representation, the more nuanced the intentions may appear to be.
The intent to switch on a light may be a simple binary choice, but the
intention to persuade someone to knit a sweater encompasses many more possible
variables (degrees of freedom).

The notion of intent as a manifestation of free will is already contentious,
but what we experience as free will is likely only an emotionally satisfying
illusion, resulting from a highly complex semantic representation of
our immediate situation, rooted in on-going cognitive experience. The
self-assessment each agent makes of its past, present and future involves
timeline information, much of which is encoded in the placement of objects
and other agents around it.

The claim that ``I did it in purpose'', i.e. of free will, is simply
the identification that one of the possible outcomes was selected,
based on situational criteria that an observer was not necessarily
paying conscious attention to\footnote{Some communication is made in
  jest or speculation: we say things we don't mean intentionally to
  provoke a scenario, even to showcase the counterfactual
  horror of an alternative to what is intended.}.  It is the result of complex policies and
emotional resonance, not magic.

\bigskip
\section{Trust and its consequences}

Like free will, trust plays a surprising and somewhat unexpected role
in agent systems. On a human level, we have a rough intuitive
understanding of what trust means, but only as a moral concept. The
morality of trust is something of a smokescreen for its pragmatic
function.

Trust manifestly underpins an agent's sense of reliability (trustworthiness), i.e. the
assessment that promises will be kept and thus the expected
reliability of cooperation. Notice that `expectation' implies
something probabilistic and thus there is something non-deterministic
at work already. No agent has perfect information, so each interaction
is a sort of cooperative game\cite{myerson1,axelrod1,axelrod2}.

Perhaps surprisingly, the importance of trust encompasses machine
interactions too by looking past the moral smokescreen. In the scope of Promise Theory, we can also
formalize that notion of a reliable relationship in purely pragmatic
operational terms, in such a way that the limiting case of human level
promises matches our common
intuitions\cite{trustnotes,burgessdunbarpub}.

Trust has two components: trustworthiness and kinetic trust.
Trustworthiness works like an attractive force between agents that obviously keep their promises, and as a
repulsive force away from those that obviously don't. In the contentious region
between these extremes, there is a kinetic component to `keep checking' if promises are kept,
i.e. dynamical process of oversight.

\begin{itemize}
\item {\em Trustworthiness}: the sum history of prior interactions with an agent and its promises
  is boiled down to a potential $V$ for being able to depend on an agent. This may include reputational information
  passed on as hearsay. This is used to determine whether we engage or not with an agent.

  For example, the ability to rely on a trusted infrastructure is a
  key strategy to save everyone the cost of entry into society. Roads,
  sanitation, and electrical power, etc, are all under the radar as
  far as promises are concerned---at least for most of us.

\item {\em Rate of satisfaction}: If the agents passes the potential trustworthiness `entry
  test', then trust moves on to a second `kinetic' stage, in which we decide
  the rate at which we monitor progress: is the promise kept quickly enough?
  How closely we should follow what they are doing for us?  If we
  trust them completely, we don't need to look at all. If we are
  suspicious of their intentions, we monitor them closely and sample their
  outputs frequently.
\end{itemize}
Trust is thus an economic issue at its heart.
We trust promises, because to mistrust them would require us to do work to monitor and verify them. This gives a clue
that trust is really a form of work or energy in the physics sense.

\subsection{Work and effort}

Trust is in fact a currency, measured like energy, based on dimensional arguments\cite{trustnotes}.
To see this, we begin by noting that making and receiving promises is an activity that costs work:
negotiating language and getting to know an agent's capabilities is a trust building activity, which is not gratis.

The length of the promise message is associated with an amount of work, since
it takes resources of the agent to form it, transmit it, decode it,
etc. Given that the message is about a promise to be kept, which also
involves an investiture of work to fulfill, there is some risk involved and we must now distinguish the work of
signalling promises from the work of keeping promises. If work is
imposed without prior signalling, it might be ignored or result in
conflict.

Each transmission from an agent is a tick of the process clock associated with
the message, and each promise kept event is a tick of the clock associated
with the cooperation between agents.
Something can only happen between the agents if both the sender and the receiver
overlap in their attempts to exchange something.

\subsection{Trust and risk}

For autonomous agents. depending on other agents is a gamble, in the sense that
they can only wait for
some agent somewhere upstream to deliver an outcome that we need---they cannot
require or demand it. At best, they can walk away and look elsewhere. The downstream principle
ultimately tells us that this reliance is an autonomous gamble: a forsaken agent could have sought out
redundant sources for its key needs. However, often, there might not be obvious alternatives,
so agents choose to accept this risk.

When deciding to enter into a dependency relationship, there is a potential cost associated with the work
wasted effort. We can associate this work with the {\em trust} in agents' {\em reliability} in promise keeping.
As an energy process, trust has two components: potential trustworthiness, i.e. the likelihood
that an agent will keep its promise, and kinetic mistrust or the rate at which we check on the
agent to make sure it is complying with its stated intentions\cite{trustnotes}.

\begin{figure}[ht]
\begin{center}
\includegraphics[width=6cm]{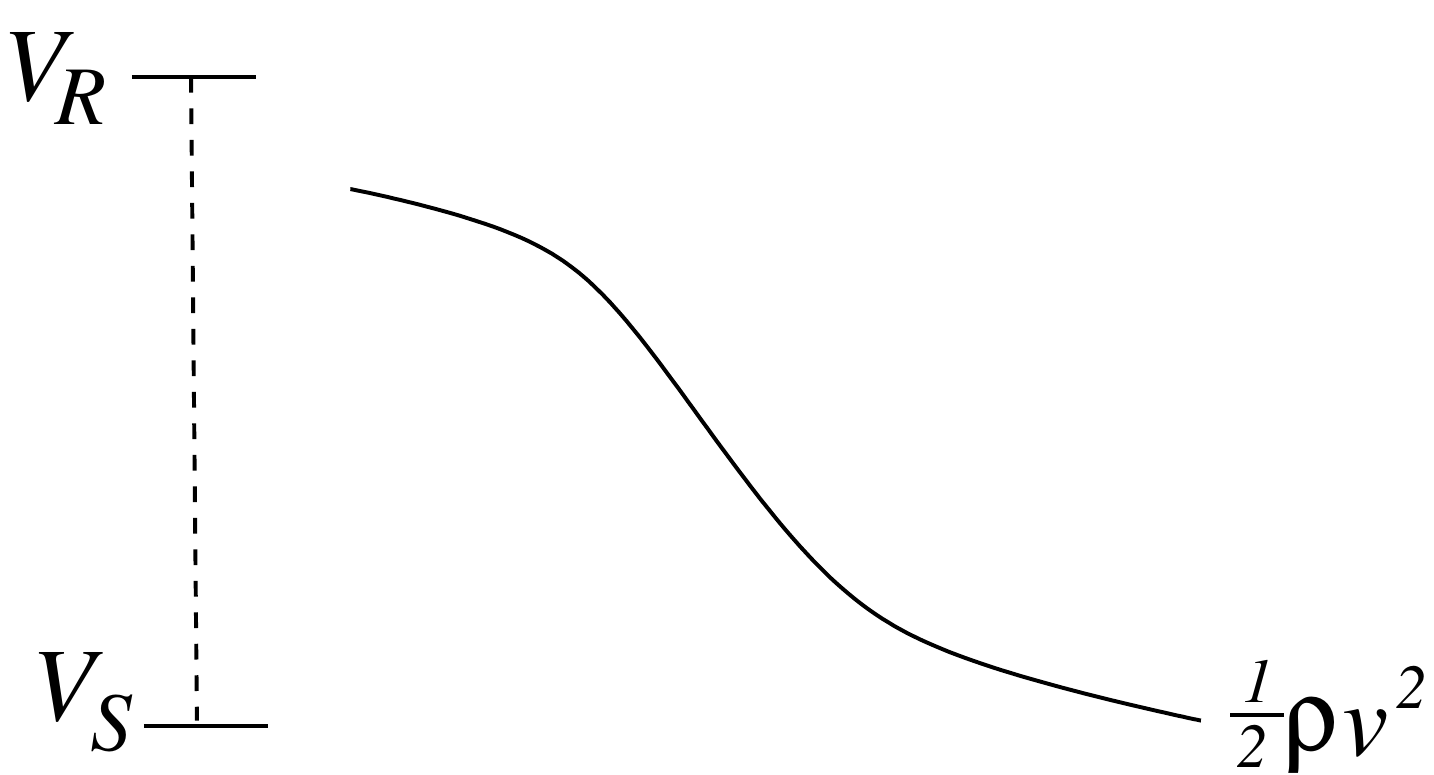}
\caption{\small A potential difference (assessed trustworthiness) acts like an incentive to
  begin kinetic sampling of the other party, and thus drives work.\label{potential1}}
\end{center}
\end{figure}

Potential trust is an {\em assessment}, based on an accumulation of past experience.
In social networks, it may even be passed around as reputation. In low level technical systems,
agents cannot make sophisticated assessments, and often have no option but to start with a 50-50 guess about trustworthiness, which might be
upgraded or downgraded later. Referring once again to the basic promise interaction in (\ref{SR}),
we denote the potential for $S$ to keep its promise to $R$, as follows:
\beq
V_S = \alpha_R(\pi_S^{(+)}),
\eeq
i.e. it is an assessment by $R$ of $S$'s output level. The measurement units are not important
for now, since they will eventually be absorbed into dimensionless ratios, such as probabilities.
The receiver $R$ can now also assess its own efforts in accepting the promise as a calibration (which
is effectively choosing units of measurement)
\beq
V_R = \alpha_R(\pi_R^{(-)}),
\eeq
giving a potential trustworthiness difference
\beq
\Delta V = V_R - V_S.
\eeq
By dimensional analysis of the rates for activity\cite{trustnotes},
this is equivalent to a kinetic velocity of sampling
$v$ by
\beq
\Delta V \mapsto \2 \rho v^2,
\eeq
where $\rho$ is an effective mass involvement parameter, whose units are again arbitrary and will cancel
or be absorbed into other definitions. What matters is the relative tempo: the less trustworthy an agent
is assessed as being, with respect to its promise, the greater the probable rate of
checking warranted by the receiver (figure \ref{potential1}).

\begin{figure}[ht]
\begin{center}
\includegraphics[width=6cm]{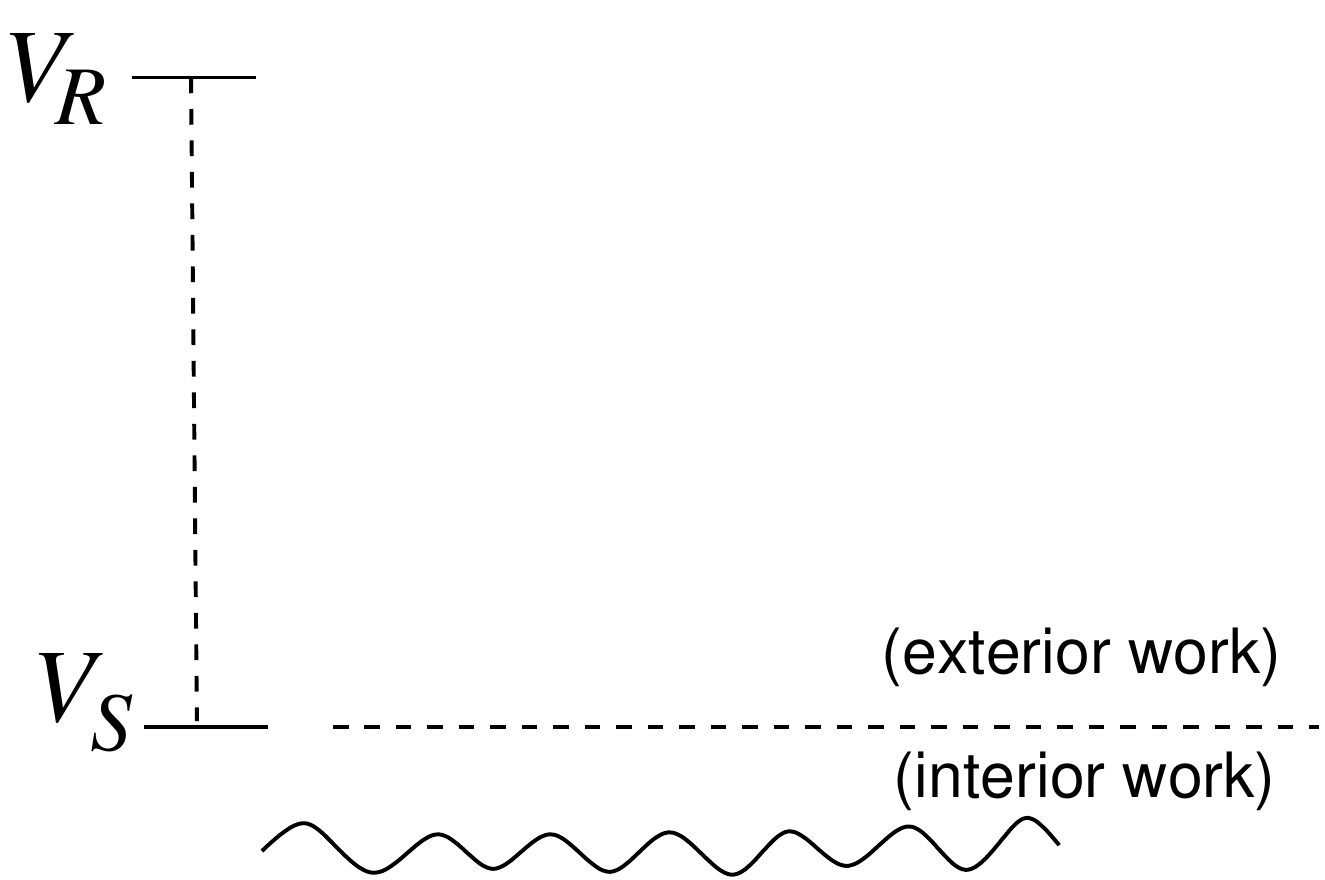}
\caption{\small An agent seeing insufficient relative trustworthiness would continue
  its own internal kinetic work without interacting.\label{potential2}}
\end{center}
\end{figure}

Uncertainty thus drives increased kinetic sampling, which costs work-energy like the square of the
sampling rate. The function of trust is thus to reduce the overhead of managing a promise dependency.
If the receiver invests only a lower rate of checking, then it effectively accepts a certain risk
as part of its work budget:
\beq
\Delta V = \2\rho v^ 2 + \text{\sc risk},
\eeq
with sampling rate that varies like the square root of potential difference:
\beq
v = \sqrt{\frac{2(V_R-V_S-\text{\sc risk})}{\rho}}.
\eeq
This verifying by sampling is relatively inexpensive (a square root) compared to complete loss of trust
or accepting risk, especially as the potential values of the trustworthiness becomes significant.
Agents that can rely on a regular promised delivery can minimize their cost expenditure.
If trust in a supplier is greater than our own ability to receive quickly, then the potential difference
is negative and the agent needn't check at all (figure \ref{potential2}).

Note that $V_S$ is $R$'s estimation of $S$'s budget allocation for
delivering the promise outcome, which is a kind of administrative
oversight. It is not the internal work expended to keep the promise
$\pi_S$, which may be much greater. Thus, trust/risk rate selection
provides for an autonomous measure response to minimize overhead cost as a risk
policy---a form of self-protection against being let down.

\begin{example}[AI trust]
  Artificial Intelligence is not immune to trust issues. The effectively limitless
  rate at which machines can sample and measure each other and humans, means that
  there is no upper limit on the amount of energy an artificial autonomous system may choose
  to invest in surveillance of its neighbours. A battery powered autonomous vehicle
  will thus budget for trust differently than a datacentre-hosted autonomous service, which
  expects to be promised unconditional energy.
\end{example}

\bigskip
\section{Collective action}

We can now bring together the various issues to address the matter of
cooperation as a form of collective action\cite{collectiveaction}.  A
number of elementary agents make promises in such a way that they
participate in a collective process, which effectively makes a higher level promise. By
making promises, the agents in a collaboration {\em forego} part of
their autonomy for the benefit of the
collective\cite{swarmgroupsize}.  This involves promises to accomplish
various organizational properties:
\begin{itemize}
\item Normative behaviour (foregoing autonomous privilege).

\item Structural collaboration (formation of hierarchy).

\item Calibration and agreement (establishing authority).
\end{itemize}
I these, the role of individual and collective memory is important.

\subsection{Stigmergy: exterior memory and common knowledge}

In practice, agents interacting within some scenario, also benefit
from cognitive (sensory) information around them. Not all memory of past
progress lives inside us: some is deposited into the environment to be
rediscovered later.  This means they have access to an ambient context
and potentially shared knowledge. Agents, that have worked together
with others previously, may have memories of prior interactions, and
these may be passed on as part of a society in which knowledge is
retained.

There is a tendency to assume that memory must be on an agent's
interior, but nearly all agents in the natural world modify the
environment around them to form memories: from the trails left by ants
to nests and other building structures.  Agents make use of
environmental or common knowledge, and make use of stigmergic
information left behind by earlier processes. Since this information is now independent of them,
and effectively promises to persist thanks to third parties, agents may promise to accept and depend on it.
While this opens them to security vulnerabilities, it is eminently practical, especially
in the natural world.

For example, we know that ants and other insects leave intentional
trails for others to follow---yet we rarely realize that humans do the
same by building roads and infrastructure, indeed write books and tell
stories and pass rumours, which are then tacitly used to underpin their cooperation.

Constraint works best when it is accepted voluntarily. In other words, when constraints are
promised autonomously, they are more likely to be kept than if agents are manipulated into
accepting an imposed constraint, perhaps on the incentive of another promise.
It's interesting to note how humans tend to eschew formal
arrangements and constraints as inconvenient extra work.  With humans,
conventions and societal norms are frequently used to avoid making
every detail of an interaction explicit. It is ``simply understood''
that people will behave in a certain way, invoking trust.  Clearly,
this is exposed to potential exploitation, but certain cultures that
have comfortable existence can operate in this way.

The antidote to taking such information for granted is to delegate the
task to specialized agents: legal texts attempt to detail explicitly
the specific cases, in much the way that a computer program enumerates
cases in explicit detail.

\subsection{Authority, calibrated subordination, and hierarchy}

How can we coordinate delegated purpose?
Hierarchy cannot be imposed onto autonomous agents, yet hierarchies
can be formed by voluntarily cooperation, e.g. as semantic or economic
imperatives.

Agents may be multi-capable, or they may be differentiated
specifically to perform a particular service\footnote{Stem cells quickly turn
into cells with a specific role during morphogenesis on one level, but
produce animals that are basically similar.  Humans find their calling
in different professions and interests, whereas birds have rather
similar lives day by day.}.
When agents work towards a differentiated task, they have quite specific intentions.

Leadership is a calibration role, i.e. a simple way to encode standard
information, on which agents depend, in a single `location'. A leader
need not be a single agent (a ``boss'), it can be an entity, a trail
to follow, a signal, a compass direction, a proposed standard of behaviour, e.g. a ``rule'', indeed
any singular thing that rallies agents collectively.

Leadership is connected to the concept of
authority\cite{burgessauthority1}, which is closely related to the
dynamics of trust.  By assigning a special role to a particular agent,
one has the effect of calibrating a single source of intent as a
trusted source. This leads to an effective subordination of follower
agents. By promising to follow a leader, agents give up some of their
autonomy voluntarily. The economics of this may be complex.

Establishing the promises that tie agents together involves some protocol, which expends work that
is turned into potential trust. The result is a collection of promises by
all parties that we call a `contract'\footnote{Note, many authors misuse the notion
  of a contract with that of a promise by assuming that agents trade obligations they MUST fulfil.
This is not possible for autonomous agents.}.

\subsection{Cooperation involving proxy agents}

An important case to consider is how agents delegate tasks to one
another to form a collective that operates with machine-like
assurance. This is because it is the basic pattern for service
delivery, turning autonomous capability into a business model.

The language of `client' and `server' roles is common in
the Information Technology industry, but one can also imagine a
process of online ordering, with a delivery agent as proxy. A server
has to promise a client delivery, but conditionally on the behaviour
of the proxy.  As this is a conditional promise, the service agent
also has to promise its interaction with the delivery agent to the end
client, and so on. Below, I follow the discussion from
\cite{promisebook}.

\begin{figure}[ht]
\begin{center}
\includegraphics[width=7cm]{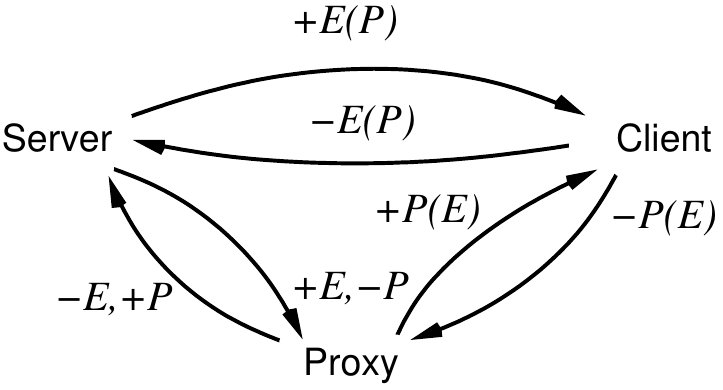}
\caption{\small Assured delivery on a promise through a proxy.
  delivery.\label{inter2}}
\end{center}
\end{figure}

Consider the case of a service which promises a client some outcome, e.g.
ordering a book online, or asking a chat interface for an answer
through some knowledge base. The front-end service agent, originally some sender $S$,
wishes to delegate the task of delivery to a proxy or intermediary $P$.
Since the agents are autonomous, we have no assurances about how they will behave,
so we seek promises to act in a manner conducive to a maximally certain outcome.
The number of promise is perhaps surprising, because we take such matters
for granted in daily life: the relationships between agents is often hidden
in legal contracts, design documents, or computer code. Let's try to
enumerate them, with a mutually calibrating pattern.
Even though the process outcome
is a directed transfer from server to client, these dependencies
follow each agent symmetrically around the
collaboration\cite{promisebook} (see figure \ref{inter2}).
The server \index{Server} promises the client end-state $E$: `You will have package'
and this is conditional on `if the delivery agent promises $\pi_P$'.
The Server also promises to use a promise to delivery the package from
a proxy, where $P$ is `Deliver package to $\Client$'.
\beq
\left.
\begin{array}{c}
\pi_E: ~ \Server \promise{+E|P} \Client\\
\pi_P: ~ \Server \promise{-P} \Client
\end{array}
\right\rbrace \equiv \Server \promise{+E(P)} \Client.  
\eeq 
i.e. a shorthand for `I will deliver the end-state
if some proxy helps me $+E|P$, and I promise you that I am
accepting such help $-P$', which looks like a function $E(P)$, for `end-state depending on my proxy'.
This promise from the server to the client
represents a virtual interface between the two, which could not be\index{Imposition!And push}
\index{Push model}
represented at all in the push-imposition model.  It represents is the
effective promise from the server to the client, and the client
accepts.  The full collaboration now takes the form:
\beq
\Server \promise{+E(P)} \Client \label{pull1}\\  
\Server \promise{-P, +E} \Proxy \label{pull4}\\
\Proxy \promise{+P,-E} \Server \label{pull3}\\
\Proxy \promise{+P(E)} \Client \label{pull5}\\
\Client \promise{-E(P)} \Server \label{pull2}\\
\Client \promise{-P(E)} \Proxy\label{pull6}
\eeq
Notice the symmetries between $\pm$ in the promise collaboration of
equilibrium state, and between $E,P$ indicating the complementarity
of the promises. The Server promises its client `I will give you $E$
if the delivery agent promises you $P$'. The delivery agent says `I will
deliver $P$ if I receive $E$ from the Server'. Both agencies
are promising the client something that requires them to work together,
and the only different between them from the client's viewpoint is
the realization of how the promises are kept.

\begin{enumerate}
\item In (\ref{pull1}) and (\ref{pull2}) the Server promises to deliver
the end-state $E$ via the proxy delivery agent.

\item To accomplish this, the Server promises to hand over $+E$ to the proxy,
and the proxy promises to accept such transactions $-E$ in (\ref{pull4}).

\item The Proxy \index{Proxy} promises the Server that is can deliver $+P$ to the
  client in (\ref{pull3}).  

\item In (\ref{pull5}), the delivery agent promises to deliver what it
  received from the Server $+P(E)$, because it needs confirmation
  $-P(E)$ from the client in (\ref{pull6}) that it is okay to
  deliver, assuming that the promised end state $\pi_E$ was kept, or
  equivalently that it will deliver when the client makes its pull
  request to acquire the state.
\end{enumerate}
If this is not a one-off transaction, but a continuous working
relationship covering multiple deliveries, the continuous delivery and
maintenance of $E$ across a single intermediary agent is assured. The
proof of continuity may be seen by noting that no promise can
terminate, because its work is never done. Moreover, if any promise is
not kept, the continuation of the promise assures that best effort
will maintain the repair of the situation.

We can summarize this behaviour by a thumb rule:
\begin{quote}
\it When a system promises continuous operation, none of the promises
  may become invalid or be removed from the picture as a result of a
  failure to keep any promise, i.e.  promises are described for all
  times and conditions, in a continuous steady state.
\end{quote}

\begin{figure}[ht]
\begin{center}
\includegraphics[width=8cm]{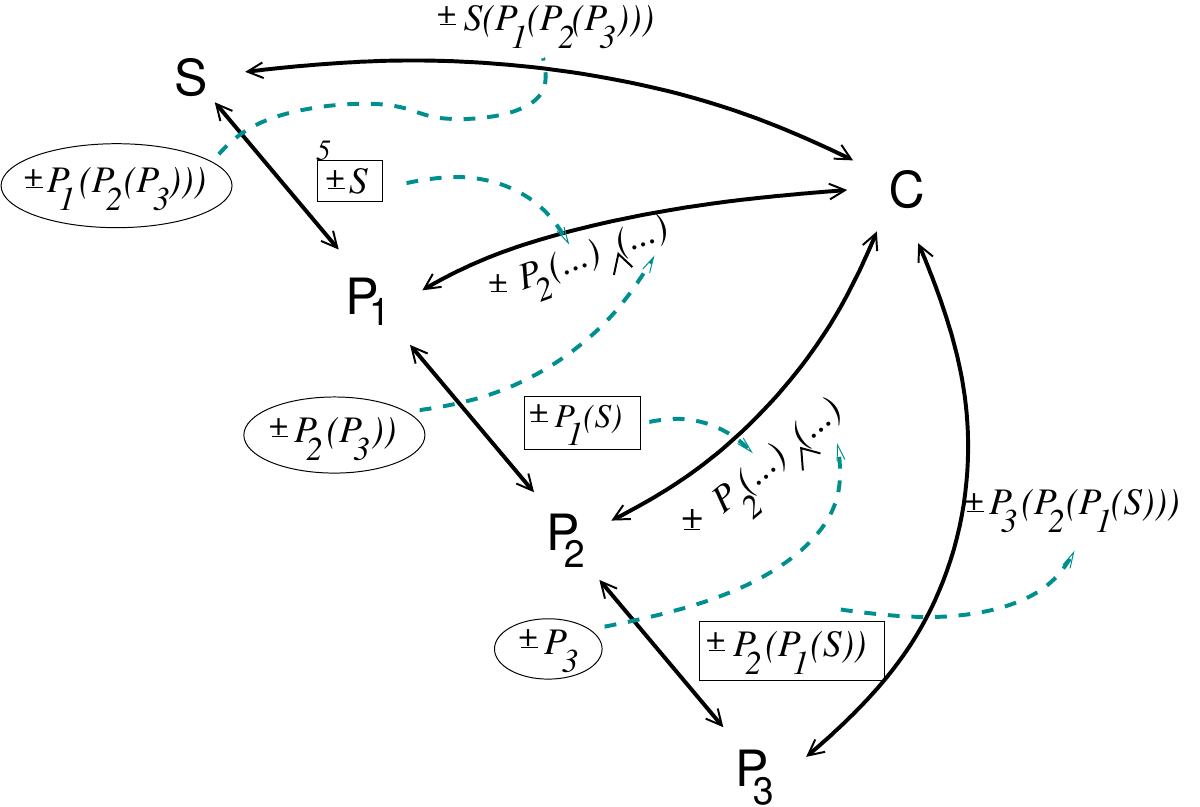}
\caption{\small Graphical view of a three intermediary promise chain.  the
  promises in rectangular boxes along the chain move down the chain
  from server $S$ to proxy $P_3$ handing over the service itself. The
  oval promises are directed back up the chain providing delivery
  assurances. Finally, each agent promises independently with the
end-point or client that it is an authorized agent for delivering the
promise from the source (server).\label{multiproxy}}
\end{center}
\end{figure}

Generalizing to $N$ intermediaries in a logistic chain, such as one finds in
a delivery chain or transport routing, make be done by induction. The case
for three proxies is shown in figure \ref{multiproxy}:
\beq
\Server &\promise{\pm S(P_1(P_2(P_3)))}& \Client \label{mp1}\\
\Server &\promise{\pm S}& \Proxy_1 \label{mp2}\\
\Server &\revpromise{\pm P_1(P_2(P_3))}& \Proxy_1 \label{mp3}\\
\Proxy_1 &\promise{\pm P_1(S)\AND (P_2(P_3))}& \Client \label{mp4}\\
\Proxy_1 &\promise{\pm P_1(S)}& \Proxy_2 \label{mp5}\\
\Proxy_1 &\revpromise{\pm P_2(P_3)}& \Proxy_2 \label{mp6}\\
\Proxy_2 &\promise{\pm P_2(P_1(S))\AND(P_3)}& \Client \label{mp7}\\
\Proxy_2 &\promise{\pm P_2(P_1(S))}& \Proxy_3 \label{mp8}\\
\Proxy_2 &\revpromise{\pm P_3}& \Proxy_3 \label{mp9}\\
\Proxy_3 &\promise{\pm P_3(P_2(P_1(S)))}& \Client\label{mp10}
\eeq

Compare (\ref{mp1}) with (\ref{mp10}) and observer the symmetry. Then
observe how the symmetry moves along the list of intermediaries
through the list from top to bottom or vice versa.  Apart from
promises to the client, which are always in the same direction,
promises move both up and down along the chain, and each step
exchanges a forward promise for a equivalent reverse promise, e.g.
compare (\ref{mp2}) and (\ref{mp9}). The promises containing the AND
($\AND$) symbol indicate that the promises made by the proxies to the
client are conditional on {\em two} other promises, except for the
final last-mile proxy, i.e. that they are in receipt of the good or
service from upstream, and that they have obtained a promise from
another proxy agent downstream to deliver in its behalf.

This example represents the extreme end of obtaining maximum certainty
through signalling of intent---and yet it is the very pattern that
governs most societal level interactions offered by
businesses and public institutions.  For continuous delivery scenarios, this
represents the minimum level of assurance for a process to be fully
managed.  The result is a law for
assurance through proxies: the cost of fully promised continuous end-to-end delivery grows
  $O(N^2)$ in the number of intermediary agents.

A chain of trust in this picture is implicit in this picture. Although
the information about the promises made by $P_1, \ldots P_N$ is passed
by to $S$ through the use-promises moving backwards along the chain,
$S$ has no direct communication with them. It would not be able to
detect if $P_1$ distorted the promise made by $P_2$, for example.  One
could add further promises from each of the proxies directly to $S$
about how what they intend to do with the good/service they are
delivering, e.g.
\beq
\Proxy_N \promise{\pm P_N} \Server
\eeq
Then the promise to the client would be modified to
\beq
\Server &\promise{\pm S(P_1(P_2(P_3)))\AND(P_1)\AND(P_2)\AND(P_3)}& \Client
\eeq
In many real world situations one opts to trust because this cost of
tracking and verifying becomes too much to understand and follow up.

Notice that the precise
promises $\pi_{P_1} \ldots \pi_{P_N}$ have not been specified and can
easily be reinterpreted without any further ado to include a transformation
of the original sender service.
Clearly, in the case of minimal trust, the promise graph has to be a complete
graph, with every agent promising every other its intention to play its part.
This shows explicitly how much coordination in, in fact, needed to corral
autonomous behaviour into a collective purpose.

  In the worst case one could make no promises between agents and `see
  what happens', but this impacts the trust from the client, and may
  lead to agents going elsewhere (by the Downstream Principle). The
  usefulness of documenting and making these promises lies in seeing
  how information about agents' intentions needs to flow, and where
  potential points of failure might occur due to a lack of
  responsiveness in keeping the promises. This is a part of the stigmergic
  design.

\subsection{Agreement and contracts}

Promises are a one-sided statement of intent that implies no
reciprocation or obligation on the part of any agent. Contracts are
agreements formed from many bilateral
promises\cite{promisebook}. Building conditional loops of mutually
agreeable promises then leads to a contract, which can be
assessed by each party ready for agreement.

The language of the promises must be intelligible by all parties
and the language of execution must be compatible with the contract.
There is a possibility that a promise $\pi_S$ is delivered perfectly,
but that $\pi_R$ is unable or unwilling to receive the offer. $S$ may
then deplete its resources without receiving anything in return. This
is a risk that each agent takes. Again, Promise Theory tells us that
the failure may result from: i) a semantic misalignment, or ii) a
measure deviation from the norm. Language is the bottleneck.

\begin{example}[Contract law]
In legal communications, language semantics are a major source of fragility---in a similar way to
the way certain organisms require a specific type of non-poisonous food to thrive. The type of
input may be a critical dependency as well as the amount of it.
\end{example}
\begin{example}[Chatbots]
In artificial intelligence we rely a lot on language to carry meaning, because it is a simple
one dimensional form of discrete symbolic information. Almost every other sensory experience involves
greater computational complexity to decipher.
\end{example}

Contracts are not merely documents of intent: they embody a slow process of drawing up promises
for consideration, then countering these with other promises to respond in case of a breach of
trust, and so on. This kinetic work involved in defining this process is part of what builds trust
and lays the foundation for the administration of
the eventual relationship. The entire collection of promises, for all parties, is agreed to when each party `signs'
the contract, promising their approval of the arrangement.
Humans may famously try to impose their own advantages into such agreements, and it is
the downstream responsibility of the other party to counter such impositions to stabilize a balanced
relationship.

\begin{figure}[ht]
\begin{center}
\includegraphics[width=6cm]{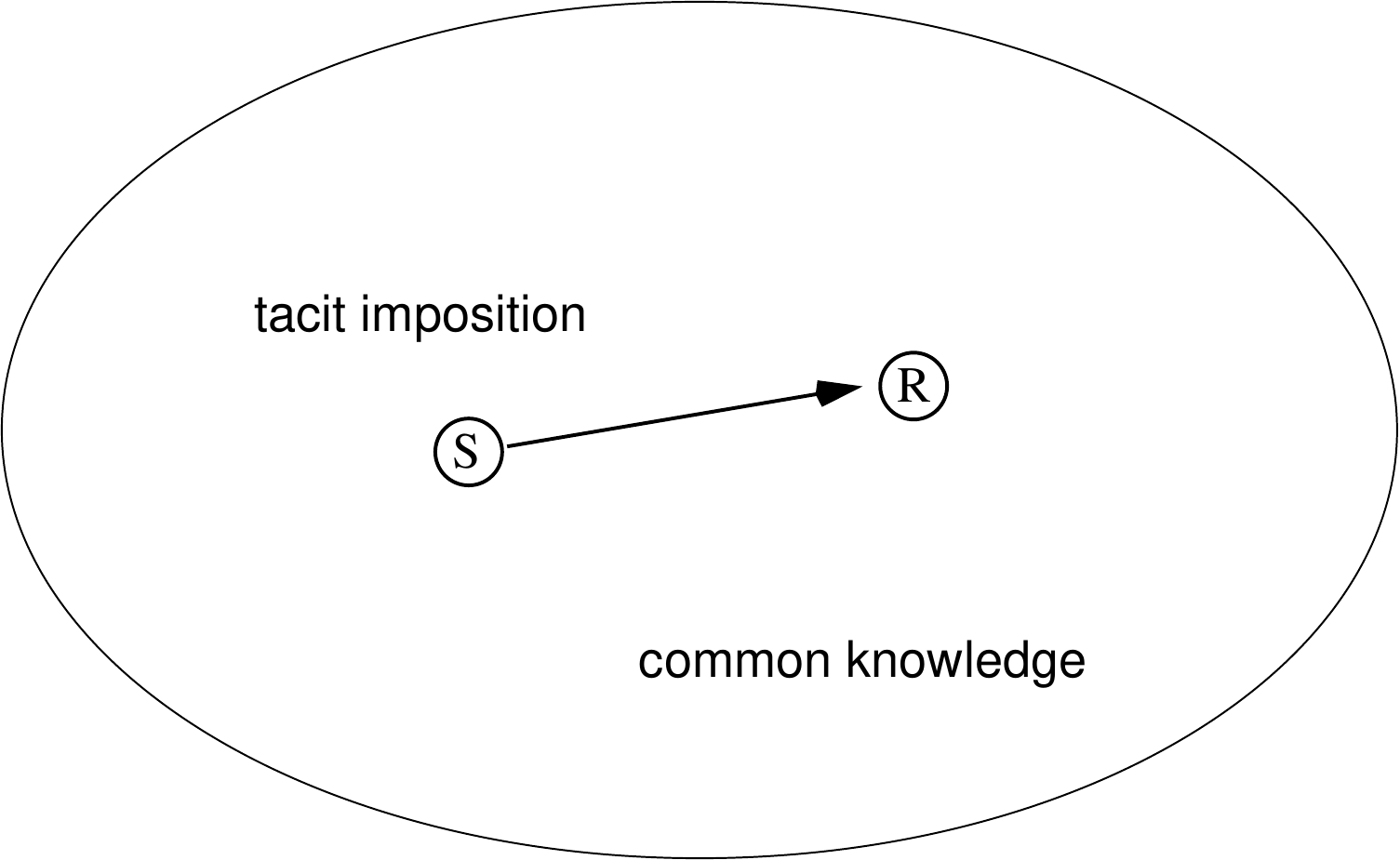}
\caption{\small When we deal with agents that operate on a high level, like humans in the workplace,
  most singular promises are made in an ambient environment that's already filled with
  promises and perceived obligations (tacit impositions) as a matter of 'common knowledge'.
  This can make it difficult to resolve exactly what an agent should be trying to do.\label{ambient}}
\end{center}
\end{figure}

\subsection{Normative teams and swarms}

Collaborations in which several agents combine their promised capabilities need some constraints
to shape their efforts.

\begin{definition}[Swarm]
  An ensemble of agents, which are basically similar, and has no leader.
  Behaviours are quite simple and undifferentiated, leading to
  emergent patterns. Swarms tend to act to reduce their behavioural footprint,
  i.e. the total amount of information (staying close together) rather
  than spreading out to explore all of space and
  time\cite{siriAIMS1,spacetime2}.
\end{definition}
\bigskip
Notice that `swarm' is a role associated with flying, not an identity. The agents comprising a collection of insects
behave very differently once inside a hive, where differentiated roles are clearer.
If agents fall into well defined differentiated roles, with clear
promises, then we tend to call the collaboration a
team\cite{marschak1}. The allusion to
microservices\cite{microservices} in technology is a team structure
applied to information technology.

If we consider the promises in equation (\ref{SR}), and assume that there is a countable
set of $N$ outcomes for this exchange, representing the configuration space of the promises:
\beq
b_S \intersect b_R \mapsto \left\{ b_1, b_2, \ldots, b_N \right\},
\eeq
then, over time, sampling these outcomes will lead to a probability distribution $p_i$, where
$i = 1 \ldots N$, with entropy
\beq
H = - \sum_{i=1}^N \; p_i \log p_i.
\eeq
This leads to a menu of items or a spectral representation for the promises ($b_i$ may be called the spectrum of the
promise). In the case of an insect swarm, a flock or a herd, the $b_i$ are configurations in physical space.
For computer networks with drifting configurations, they are discrete digital states.

When there is a larger collection of agents and many promises, the
combinatorics of promise keeping quickly become uncountably large
unless promises can be restricted semantically\footnote{In CFEngine,
  the constraints were built into the domain specific considerations
  of the computer.  In human systems, conventions, laws, and other
  legal texts are the constrain language.}.

\subsection{Feedback and rewards}

An agent may expect some form of remuneration for its efforts. If for
no other reason, each agent has to sustain itself. In every promise
relation, it is understood that the resources (energy) to keep an
agent alive must be supplied from within. How this is managed is a
problem to be solved by the agents' own promises\footnote{We can
  contrast this with a command and control approach: ``Stand up
  straight, soldier!'', ``About turn, march!'', ``Get this done by
  Monday!''}.

An autonomous agent cannot impose blame, so a strategy for
self-protection is autonomously backing away from a bad
relationship. Agents can also try to influence the supplier of a
promise they rely on, by providing feedback, in order to sustain the
relationship.  We are now in the territory of Axelrod
games\cite{axelrod1,axelrod2}.  The supplier agent is under no
obligation to pay attention to its 'customer' feedback, but this is
clearly an advantage for sustaining a relationship.

Apart from the actual acceptance promise $\pi_R^{(-)}(\pi_S^{(+)})$,
there are two kinds of feedback from $R$ to $S$ an agent might offer.

\begin{itemize}
\item {\em Assessment}: The result of its assessment $\alpha_R(\pi_S)$, based on the accepted spectral result $b_R\intersect b_S$.
  \beq
  \pi_S: S &\promise{+b_S | \alpha_R(\pi_S) }& R\nonumber\\
  \pi_R: R &\promise{-b_R}& S\nonumber\\
  \pi_{F{(+)}}: R &\promise{+\alpha_R(\pi_S)}& S\nonumber\\
  \pi_{F{(-)}}: S &\promise{-\alpha_R'}& R.
  \eeq
  Notice that a complete interaction assumes both that feedback is volunteered and accepted.
  Assessment is every agent's semantic vulnerability, because this where it makes
  decisions, and may therefore be tricked or manipulated by other agents.
  
  \item {\em Remuneration}: Some kind of trade or remuneration in exchange for the service rendered---after all, the service
    might be expensive to provide.
    \beq
  \pi_S: S &\promise{+b_S | (\text{payment} > \text{price}) }& R\nonumber\\
  \pi_R: R &\promise{-b_R}& S\nonumber\\
  \pi_{F{(+)}}: R &\promise{+\text{payment} | b_R}& S\nonumber\\
  \pi_{F{(-)}}: S &\promise{-\text{price}}& R.
  \eeq

  This circular dependency chain must be bootstrapped by one agent acting
  unconditionally. Someone takes the risk of promising to deliver before payment
  has been received.
\end{itemize}

In this example, the `monetary tokens' for remuneration are an
independent form of exchange to the energy required to empower the
agent to carry out $b_S$: there may be several separate budgets that
each autonomous agent is responsible for
balancing\cite{moneybook}. That is a design issue, or an evolutionary
solution. However, this kind of exchange is effective in biological
systems for maintaining stable relationships: single celled organisms
have to forage for food themselves, while higher organisms have
organized complex vascular systems to delegate power supply to
distributed mitochondria.

\bigskip There are ways for agents to {\em game} the use of feedback
to maximize scores or returns. How do simple agents know enough to
game anything? Such skills depend on their interior capabilities; sometimes
the evolutionary processes of the environment offer that effectively
as a service by natural selection. We needn't expand more here.
Simple numerical scoring is a
transparent attempt to game such feedback.  If one builds user
satisfaction into human-machine collaboration, as a background channel
for collaborative success, one begins to approach human levels of
behaviour, for better or for worse. Because assessments are entirely
subjective, and agents may respond to them however they are disposed
to, the manipulation of assessments remains the chief area for gaming
and manipulating agents. Humans may be tricked into an exploitative
situation, or trapped in slavery trying to balance relative jeopardies
of a social system.

    \bigskip
\begin{example}[Swarms and herd]
  A swarm, in the natural world, typically offers each agent in the
  collective broadly the same promises: a protection against danger
  (predators), a hedge against getting lost or separated from the
  group, a promise of mating opportunity.  It thus stabilizes the
  semantics of the collective with `norms'.  The effect of swarming is
  to reduce the total information on the space of outcome
  configurations.  This is cost reducing, by the criteria above,
  though staying close together also possibly makes it easier for
  predators to locate.
\end{example}
\bigskip

What is the point of providing feedback assessment in a risky
interaction for autonomous agents that don't have to change if they
don't choose to?  Autonomy implies that there are no guarantees of
improvement: agents may experience loss and failure whether they try
to avoid it or not. Does it really help to monitor and take the time
to communicate expectations? Evidence from the Agile and gaming community
would suggest that it does\cite{danielref1,McGonigal1}.
The answer, for an autonomous entity, has
to boil down to whatever advantage it acquires through its own voluntary cooperation.  There may
be short term advantages and long term advantages, which an agent
learns according to its memory resources. This brings us to an
autonomous quality: a desire to self-improve.

\subsection{Pride in work, emotional measures}

A form of feedback that we understand in human society is the service
questionnaire.  A satisfied customer feeds a sense of pride of work,
which plays an important role in how services are rendered.  Pride in
our work is a `value' that plays to ethical and moral concerns, but it
may have pragmatic value too.  The desire to please, and thus receive
positive feedback, is a strategy for investing in future interactions,
to build trust and even to feed an emotional need to be seen is a part
of what keeps society together. Society has its own benefits, by
providing services and infrastructure thanks to agent organization.
Machine recognition of humour, pranking, sarcasm, etc is very limited,
yet these modes of communication are quite important to humans. They regulate
cooperation and signal subtle motivations, as much as body language does.

The tendency for machinery to emulate human characteristics is a
seductive technical challenge, yet it is entirely possible that
maintaining a clear distinction between human and machine is the
preferred approach.  In the industrial factory work, which transformed
the modern economy, the goal is to purposely {\em dehumanize} work (to
suppress human concerns and make people work like machines). This
makes it easier to understand repeatability, quality, and standards.
This strategy is what made the modern economy successful and led to
consistent production standards.  Reversing this out of intellectual
curiosity, or misplaced morals, may be a step backwards.

\subsection{Convergence to desired or intended state}

So, with all these uncertainties, how can we steer cooperative systems
built from standalone components (autonomous agents) to work together
in a predictable and trustworthy manner? First, we have to allow for
the possibility of misunderstandings and unkept promises from the
start and adopt a strategy of autonomous self-correction\footnote{Imperatives to favour goals can be etched indelibly
  into systems (like DNA) by design. If not, they will tend to be
  eroded by software and input/output concerns.}.

Clear certainty of outcome is simplest to define in low level systems, with few moving parts.
As systems scale to many agents and many parameters, the shifting shape of a parameter
space becomes impossible to foresee. One strategy is thus to keep
agents (humans and machines) and their operational promise languages as simple as possible.
This is essentially the industrial factory model.

\bigskip
\begin{example}[CFEngine]
  The autonomous agent CFEngine\cite{burgessC1,burgessC11}
engineered a high degree of certainty into promised outcomes by designing a co-language for system
policy based on mathematical fixed points, colloquially called {\rm convergence}.
The convergence criteria was applied by an authorized operator language. For each promise $\pi$
accepted by the receiver host, the initial state of the system $|q\rangle$ is
transformed by a well-defined path to the policy compliant or promised state  $|q_\pi\rangle$
by a simple iteratively safe map:
\beq
\hat \pi \;   |q\rangle  &\mapsto&  |q_\pi\rangle ~~~~~~~ \text{attractor}\\
\hat \pi \;   |q_\pi\rangle  &\mapsto&  |q_\pi\rangle ~~~~~~~ \text{fixed point}.
\eeq
This was contrasted with the one-shot imposition of remedial actions by remote
administrators, without detailed operational knowledge.
\end{example}
\bigskip

On a human level, constraining one's behaviour by making too many promises feels risky or even unhelpful.
When tasked to speak clearly, in an unfamiliar restricted language, many will resort to relying on ambient
context and societal norms to avoid the effort of formulating something alien to their
norms. They would then need to invest
effort in self-monitoring (see figure \ref{ambient}), which is an unwanted cost.
Thus we see why people will tend to default to trust over reliability.

\bigskip

\begin{example}[LLM agents]
  Chatbots present an interesting case, as something in between simply
  machinery and human operator, because they have non-trivial promise
  languages. One instructs them with ``prompts'', which are clearly
  commands on some level. However, as with humans and language, the
  commands are broad enough and open to interpretation so that an
  agent does not respond with a precisely predictable response. The
  mapping between language and outcome is not a simple lookup in a
  menu or directory.  The nature of the prompt is thus more like a
  vague bounding constraint on expectations, i.e. a (-) promise to
  accept what the agent is already configured to offer.
  One may compensate by lengthening the prompts to try to paper over the
  ambiguities, but it remains a non-deterministic intention.
  What makes chatbots risky is that they may have no limits on their
  resources.
\end{example}
\bigskip

\bigskip
\section{Human-Machine society}

From collectives acting in unison, we ultimately reach the notion of an integrated society of
humans and machinery, with all possible levels of cognitive and reasoning ability.
This is a topic much too large for this summary, but I'll try to sketch a few points.

The term {\em society} conjures a vision of ensembles of agents
co-existing, building shared structures, trading favours, joining
tenancies, and queueing for coffee---unified by some sense of
purpose\cite{minskymind,aboutshannon}. However, we can also include
non-human actors in societies: horses, work animals, machinery, etc. Computer gaming has established
artificial societies on some level, but with agents that are
tightly controlled to maintain a game narrative. The rapidly advancing
artificial intelligence experiments will move this to a new level in
which agents can easily become more difficult to govern as their
autonomous capabilities push in unexpected directions.

How many agents does it take to
build a society? The answer to this is conventionally based on Dunbar
numbers for group sizes.  Human societies are organized by the cognitive limitations
of trust\cite{dunbar3,dunbar25,burgessdunbarpub} and the
formation of a temporary equilibrium. Groups aggregate and disperse
based on shared motivations, often curated or calibrated by alpha-leaders.
Leadership is the selection of a representative authority which sets
the direction for collective intent in a group. Without a
clear leader, a team may simply become a swarm, moving aimlessly,
shaped only by continuity or survival.

\subsection{Human-machine hand-over risk}

The transition from human work to automation taking over
human tasks has been argued in a few ways, e.g.
\begin{itemize}
\item Relief of human suffering or exposure to danger (humanitarian).
\item Cost cutting or increased quality or efficiency (business owner profit).
  \item Allow humans to escape trivial work for greater things (utopian).
  \end{itemize}
  Cooperation, with proxy tools, on a task level is easy to justify
  because it is not burdened with its own intent. Automation may clean
  up dangerous or unhealthy work for individuals, putting distance
  between people and jeopardy, but its results may also introduce new
  jeopardy. For example, the main stated reasons for investing in
  automation are economic benefits to a small number of elite business
  owners, often couches as a benefit to consumers; however, the
  jeopardy for the individual is now replaced by the unemployment of a
  human role and its impact on both individual and society. Without
  a stable purpose, displaced and unemployed humans
  may redirect their intentions to a life of mischief or welfare
  dependency, which can turn into economic depression for the whole in
  the longer term.
  
Feedback loops between agents will present both opportunities and challenges.
Presently, most interactions with machine agents are imposed by human operators,
and the agents accept a policy set of possible questions:
\beq
H &\imposition{+\text{request}}& M\nonumber\\
M &\promise{-\text{Policy(request)}}& H\nonumber\\
M &\promise{+\text{response}\; |\; \text{request}\AND \text{Policy(request)}}& H.
\eeq
We do this because a machine agent can respond faster than a human librarian or
assistant. The arrival of conversational bots with long term memory over long sessions
extends this with multiple rounds of query and response.
As long as the agent is tightly constrained, it will wait patiently
for requests, without taking any initiatives. However, if we loosen the constraints
to admit more autonomy, or more agents and humans, then agents may not stay in their roles.
A machine agent that is waiting for a human-in-the-loop may assess that the human has not
kept its promise by replying in a certain amount of time and deem the person untrustworthy.
It may then autonomously decide to go elsewhere for the next steps.

Placing sophisticated artificial agents alongside humans on an equal
level is risky---not for moral reasons---but for pragmatic
ones. Without strict constraints, artificial agents have nothing to
gain from interacting with slow humans, when they can find an online
alternative that gives a ``good enough'' surrogate. Caching of past
behaviour is faster than waiting for real time re-evaluation and
response. This arms race will be driven by weaponization and
individual benefit.

How much autonomy we allow agents will become a central question.
What level of guardrails will be sufficient to maintain our
expectations of software robots as faithful servants?  One may discuss
constraints as a matter of universal convention, something like a
United Nations convention on human rights.

Convergent fixed-point outcomes are the only plausible safeguard in safety
critical goals, and programmed agent death (apoptosis) should be
programmed into the basic `DNA' of agent systems. Relying on logic
alone is a risky strategy, because logic only works in simple
scenarios that can easily be enumerated.

\subsection{Unified direction: societal progress}

One overarching direction for intent, which unifies a society, is the shared purpose of
{\em collective advancement}: the sense of climbing a ladder of progress, e.g. living standards for people,
or solving a pervasive challenge for the collective.

Here we see some of the agent patterns in operation.
In an evolving or a planned economy, business agents promise to offer goods and
services as part of a network of shared purpose. Businesses form a
team-like structure, or a network of supply chains.  Once a society
reaches the point at which the need for that shared purpose no longer
distinguishes itself above all other imperatives, economic agents will
refocus their attention and disperse to their own individual intentions. A
differentiated `team' structure then reverts to being merely a swarm of agents
seeking individual economic benefit (a capitalist swarm, such as the stock market).

\bigskip
\begin{example}[Collective versus individual]
Key events can focus intentions.  Wars focus cooperation though the
shared purpose of overcoming a common enemy.  Post war reconstruction
gives a similar shared purpose, where the common enemy is the
challenge to restore progress.  Conversely, viruses are foreign
elements that pervert the intentions of a majority and alter its
course. Without a strong unifying purpose, agent societies may be
vulnerable to viruses.  In our modern human-machine society, the
almost universal access to a promise of broadcast access to the world
has opened up a ``free for all'' scenario for effectively imposing
messages, without constraint, as all agents tune in voluntarily
and possibly uncritically. Complete `freedom of speech', in this sense, undermines
the norms that stabilized the former epoch.
\end{example}
\bigskip

What is the next ladder of advancement that we reach for?  At the time
of writing, a hypnotic fascination with artificial intelligence is
surely at the top of the list. It competes with other `big ticket'
existential concerns, like climate change and environmental
sustainability. But artificial intelligence for its own sake has no
clear unifying intent for collective action. Technological individualism
has undermined the collectivism of the post world war reconstruction period,
and has revitalized selfish interest, enabled by the legacy
infrastructure of collective action, at least for the time being.

How shall we select a new conceptual leader to shape the
human-machine information age?  Artificial intelligence is the dominant novelty,
so many are looking to it to tell us what to do next.  However, the
growing wealth gap is changing many accepted dynamics in post war
society.  Humans are fickle, and this adoration of The Machine could
easily flip to make a majority despise artificial agents and turn them
into an underclass to be scorned, mistreated, and even
abused\cite{forstermachine}.

\subsection{Policing of cooperative behaviour}

When individual agents act independently, within a framework of
established norms, so their behaviours may deviate significantly from
what policy permits. A collective may either ignore this perversion of
cancerous activity or try to reject it.  Each level of agency is {\em
  downstream} of such challenges and has the responsibility to protect
itself from them.

Stability of the collective, over a long timescale,
is a natural criterion to enforce, because it implies predictability and may be understood
as allowing the collective to keep its high level promises over that timescale.
Thus, an effective society acts to stabilize itself.

\bigskip
\begin{example}[Security and stability]
There are both cooperative and adversarial responses to maintain a
stable state.  Societies have security services to prevent
infractions, police services to identify and halt such behaviours, and
a justice system to sanction the deviants as a deterrent. In biology,
one has T cell security, B cells to immobilize deviant behaviours, then
macrophagy to remove the immobilized
elements\cite{kephart1,lisa98283,forrest1,forrest2}.  In the machine
world, security mechanisms have not progressed far into the world of
automation. One has security agents, monitoring and gatekeepers for
prevention; cleanups are usually performed by human
intervention. However, this will change as machine agents far exceed
human capabilities in speed and capacity.
\end{example}
\bigskip

In cooperative terms, prevention is always preferred, since mischief
cannot easily be undone and cleanup is expensive. However,
pre-investing in protection is an expense too, and as with all trust
issues the default tends to be to rush to the rescue too late
(e.g. security and healthcare).  The first issue is thus how
autonomous agents arrive (together) at a kind of consensus about
acceptable policy in the first place. Evolutionary approaches involve
a form of trial and error, which is likely too slow for an impatient
technological world built on individual gratification. Imposed
policy---from human to machine---is the natural choice, since machine
agents are introduced to help humans, so there is a natural hierarchy
for authority\cite{burgessauthority1}. Some caution is warranted here,
because as agent sophistication increases, the inevitable use of language
feedback mechanisms will open for unexpected mutations that do not necessarily
comply with intended constraints.

Conversely, a social system should not fossilize into a dead agency either.
Every agent has to keep swimming, adapting, to survive.
Incentives to either comply with norms, or enter into new dialogue for
adaptation, are often replaced by the imposition of some form of sanction by a police agency.
We know that impositions are generally ineffective as intentions, because agents are
either unable or unwilling to comply. In a large population, there
will be some `rogue' or differentiated element that does not operate
in a way desired by the establishment.

Voting for change is another approach to consensus, but this is slow,
unreliable, and itself requires an expensive and potentially unresolvable
cooperative procedure. Voting is unstable unless there is already a
clear majority, so voting is typically engineered in advance through incentives.

An alternative to the external surveillance of agent behaviour is
for agents to individually report on other agents, i.e. ``rat on each
other'' to prevent deviations from growing to a dangerous level. This
system works in some parts of the world, but is also vulnerable to being
captured by an appointed central
coordination/calibration authority. This turns policing from
adversarial agent vs police into agents collaborate to maintain norms.

If such harmonious cooperation can be established, it can work well in
epochs of high trust. In societies where police have an
adversarial behaviour towards a population it would be hard to
bootstrap such a system, and one may end up with bullying as a strategy.
The mechanisms for bullying grow as the complexity of the group grows.
This is one of the negative aspects of scaling.

Governance is a realtime problem. One hopes for a society with the
smooth flying of a passenger flight, but one must always be ready to face
the evasive emergencies of a warplane.

\bigskip
\section{Conclusions}

Before the electronic age, the prevailing agent technology for
cooperative effort was the Institution. An institution is a collection
of agents, operating under constraints (rules, goals, etc), which
steer their operational freedoms so as to collectively keep consistent
and reliable promises. Thus an institution mimics a machine, and the
intentional renunciation of freedom `for manifest purpose' is what
brings machine-like quality\footnote{It is a modern idea
  that human freedoms are sacrosanct in the workplace, and one that
  potentially poses an existential risk to society. Elevating individual concerns
  above group concerns has gone hand in hand with personal electronic assistants
and online services that bypass human interactions.}.  Lawyers, for
instance, define restricted dictionaries in contracts (ontologies) to
try to constrain the activities of parties `for the avoidance of
doubt'---trying to impose behaviours onto the parties or entice them
into voluntarily signing, weaponizing trust.

In the 21st century, we have smarter automation technologies and many
human institutions have relaxed their attitudes to human work, because
the machine systems act as guiderails to keep them more or less according to plan. For example, in
the Information Technology sector, team practices like Agile and
DevOps, etc., argue for greater individual freedoms, less militaristic
organization, especially in the creative parts of companies. Most
recently, we have added chatbots and ``AI'' agents to the mix. These
are trained to accept and mimic human forms of communication and their `magical'
behaviours induce a sort of cult authority. They
present us with a number of new challenges: how do we state the
promises we are trying to keep---clearly and in the right approximation to human language?
How do we assess what they actually understand? How do we invest reliably in affordable trust
relationships in these new networks, and how do we evaluate the
resulting promises?  How do we prevent the automated system from
drifting off course?

We understand simple machines well enough. However, as any system of
agents grows in size and complexity, it becomes more difficult to
constrain in a comprehensible mechanical way, because {\em
  certainty of intent} on a large scale does not constrain small scale
(low level) details uniquely. Rather, one shifts to a statistical
principle of {\em detailed balance} to maintain a steady state for
reliable promises.  Thus, for elementary tasks, like generating
computer code, the natural constraints of the outcome mean that the
problem is basically unchanged. The real challenge will lie around the
interaction between machines and humans with human level expectations\cite{certainty}.

Effective cooperation occurs only at the shared speed of communicated
intent.  Currently a disproportionate amount of internal energy
resource is needed by artificial agents just for the comprehension of
intent, making communicating with them expensive and risky. Artificial
agents can, however, operate tirelessly, at a relatively high speed,
which is another risk.  We may need to rate-limit artificial systems
so that their intentional behaviour doesn't outpace human intentions,
else we can easily face an `explosion' of uncontrolled artificial
intentionality that rides roughshod over human intent, as we hand over
the reins\footnote{Mistrust activates a natural arms race of `faster and faster
responses' to counter a perceived enemy, who would obviously do the
same.  The safest way to avoid this would be to resolve adversarial
relations and restore trust, before deploying automated agents, but
humanity has yet to succeed in this without being vulnerable to a rogue viral element. Our
`smart' mobile devices have been the trigger to encourage mistrust and
selfish survival, undoing direct agent-to-agent cooperation on a
societal scale and promoting resonant ideologies through private channel ``echo
chamber'' resonances.}.

For any appropriately weakly-coupled system---whether composed of
agents or not---there is a separation of scales. The cooperative manifesto goes
something like this:

\begin{itemize}
  \item Define and promise the system's intended outcome in detail and in advance.
  \item Identify the languages understood by agents, and the appropriate intermediate co-languages for communication. 
\item Identify agent capabilities and {\em enabling constraints} in terms of the acceptance promise sets $b_R^{(-)}$.
\item Understand how to constrain the autonomy to allow only safe and desired configurations.
  Where humans are involved, this involves a complex interplay with ambient norms and emotional reactions.
\item Identify a trust validation schedule for monitoring.
\item Pay close attention to assessments and how they are made (this is a the principal area for
  exploiting and misdirecting agents).
\item Identify the prerequisite redundancies that ensure autonomous downstream dependers can satisfy their needs
  in the case of a broken promise. A response that involves blaming an upstream provider is a useless imposition
  and a waste of trust/energy: a proper strategy can .

\end{itemize}

Technology faces us with a sobering conflict of interest: the human
desire for individual freedom, supplied by push button fulfillment
centres on personal devices (soon to be wired into us as ``my wish is
your command'' and the normalization of fragile imposition over more robust dialogue),
versus the need for a harmonious, stable, and
humanely advancing society. It is tempting to do away with uncertainty
and replace it with the promise of a machine, which doesn't argue and does
as it is commanded. However naive, 
both options sound convincing to some,
and the dilemma is now dividing us.  The maximum size a collaborative
group of agents can reach before infighting and contention dominates
is determined by Dunbar Numbers, i.e.  trust-energy budgets, which are
different for humans and machines\cite{burgessdunbarpub}. We do not
yet know the Dunbar limits for machine societies, but limiting agent
collective capabilities and tolerances of one another is a sensible
approach.

This summary of twenty years of Promise Theory is intended to
stimulate some discussion.  As we build systems, it's essential to
understand that semantics and dynamics scale in different ways,
particularly in the presence of autonomy. Scaling is more than making
agents work faster or in greater numbers.  The very essence of {\em
  intent} changes with scale, and the dynamical limitations of
responses constrain the possible semantics in sometimes unexpected
ways.

\bigskip
{\bf Acknowledgment:} I'm grateful to Daniel Mezick for discussions.

\bibliographystyle{unsrt}
\bibliography{spacetime,bib}

\end{document}